# SpaceAct: Benchmarking 3D Spatial and Temporal Reasoning via Closed-Loop Action and Active Perception

Chenglong Shi[*‡], Zekai Yuan[*‡], Yue Song[‡], Jon Pu[†], Will Wu[†]
Tencent HY
**Equal contribution †Corresponding authors ‡Work done during an internship at Tencent HY*

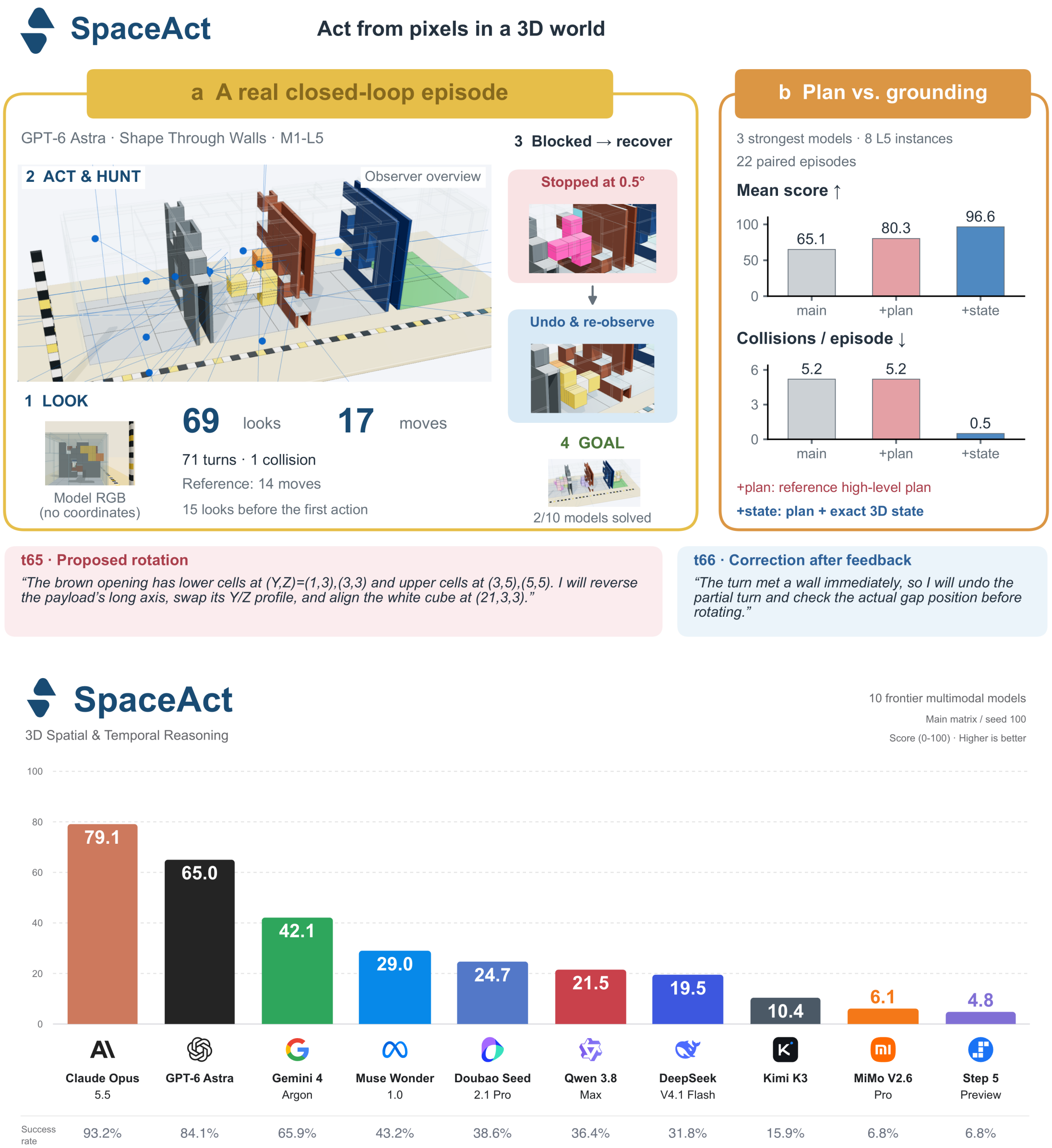

**Figure 1:** SpaceAct: acting from pixels in a 3D world. **(a)** A real closed-loop episode: GPT-6 Astra solves shape-through-walls (M1-L5) from RGB views without coordinates in 71 turns, with 69 looks, 17 moves (reference: 14), and one collision. When its rotation at turn 65 is stopped after 0.5°, it undoes the partial turn and re-observes before continuing; only 2 of 10 frontier models solve this instance. **(b)** On 8 L5 instances (three strongest models, 22 paired episodes), a reference high-level plan (+plan) raises the mean score from 65.1 to 80.3 but leaves collisions at 5.2 per episode, whereas adding exact 3D state to the plan (+state) reaches 96.6 with 0.5 collisions: even with the plan, 3D visual grounding remains the bottleneck.

## Abstract

For multimodal large language models (MLLMs) to solve problems in 3D space rather than merely describe images, they need 3D spatial and temporal reasoning: reconstructing 3D geometry and spatial relations from occluded 2D observations, anticipating physical interactions and temporal dynamics, and translating them into precise, closed-loop multi-step actions. Existing studies largely evaluate 3D spatial perception via passive visual question answering over static images or test high-level planning through discrete symbolic interactions; however, these evaluations either require no action execution or supply privileged coordinates while abstracting away collision, statics, and temporal constraints, sidestepping the core challenge of translating visual perception into continuous, precise actions. To bridge this gap, we introduce SpaceAct, an interactive benchmark that directly evaluates models' 3D spatiotemporal reasoning through closed-loop physical action and active perception, paired with SAPA, a unified Spatiotemporal Active-Probing Agent framework. Built on interlocking stud-and-tube bricks and mechanical puzzles, it comprises 108 deterministically verified instances across 36 task families in six categories, providing no privileged coordinates and incorporating controlled diagnostic modes (+plan, +state, no-camera, blind, and L7 levers). In a basic brick-bridging or aperture-threading task, for instance, the model must actively rotate the camera to resolve occlusion, read coordinates from in-scene rulers, output 6-DoF poses step by step under continuous swept-collision and calibrated statics checks, and self-correct from split-screen rollback feedback (while dynamic tasks freeze the clock during deliberation and provide five strobe frames to estimate motion periods). Evaluating 10 frontier MLLMs on the 44 hardest instances shows that scores span 4.8–79.1 (with seven models solving at most 27.3% of continuous 6-DoF tasks); for the three strongest models, a reference high-level plan (+plan) closes 44% of the remaining score gap but leaves collisions unchanged at 5.2 per episode, whereas adding exact 3D state on top of the plan (+state) closes 90% and cuts collisions to 0.5; for four further models, the plan lifts the mean score from 10.9 to 35.3 and the plan plus 3D state to 81.8; and 46% of unsolved episodes are collision or observation loops. In conclusion, SpaceAct and SAPA advance 3D spatiotemporal evaluation from passive question answering to closed-loop physical interaction, revealing that even with a correct high-level plan, current MLLMs remain bottlenecked by 3D visual grounding and closed-loop action refinement, and providing a reproducible testbed for strengthening spatial reasoning through active multi-view exploration and explicit 3D coordinate grounding.

## 1 Introduction

Completing a manipulation task in the physical world requires more than recognizing what is in an image: an agent must work out how objects move in 3D. It must infer unseen structure from partial, mutually occluding views, judge whether a rigid body will jam while rotating through a narrow passage, estimate whether a brick cantilever can carry a load, and time its release to the gap in a moving mechanism. Multimodal large language models (MLLMs) are now widely used as task planners in embodied systems, directly emitting 3D displacements, rotation angles, or assembly actions. In this architecture, high-level spatial and physical reasoning becomes a hard bottleneck: if the model flips a local tool frame, misplaces a load-bearing support by one stud, or misjudges the opening of a periodic gate by half a second, the action sequence immediately hits a wall, collapses, or misses its window, and no low-level controller can recover.

Existing benchmarks, however, rarely expose these weaknesses. The first major line of work, 3D spatial visual question answering (e.g., 3DSRBench (Ma et al., 2025a), VSI-Bench (Yang et al., 2025a), OmniSpatial (Jia et al., 2026), LEGO-Puzzles (Tang et al., 2026), and SpatialScore (Wu et al., 2026)), casts spatial reasoning as multiple-choice questions over static images or pre-recorded videos. The model is a bystander: it never moves the camera to resolve occlusion and never bears the consequences, such as collision or collapse, of a wrong action. The second line, interactive embodied evaluation (e.g., ALFWorld (Shridhar et al., 2021), VisualAgentBench (Liu et al., 2025), and EmbodiedBench (Yang et al., 2025b)), often reduces high-level household tasks to linguistic commonsense reasoning (success rates barely change when image input is removed; Yang et al., 2025b), while low-level manipulation tasks typically feed the model privileged 3D coordinates computed by an external detector and hand trajectory generation to a collision-free motion planner. Recent interactive spatial reasoning benchmarks (CubeBench (Gao et al., 2026a), SpatialAct (Liu et al., 2026), and SpatialWorld (Gao et al., 2026b)) take a step toward closed-loop evaluation, but their actions remain discrete or semantic, abstracting away continuous contact and dynamics; VABench (Zhang et al., 2026b) adds active viewpoint selection and metric end-effector commands but delegates collision checking and trajectory generation to a fixed execution layer. Third, in almost all interactive evaluations time has no bearing on the task: nothing in the scene moves on its own, so success never depends on velocities, periods, or timing. Together, these simplifications skip exactly the three core requirements of real physical manipulation: reading metric coordinates from visual scales, satisfying continuous collision and statics constraints, and acting in a world that evolves over time.

To evaluate these capabilities directly, we introduce **SpaceAct** (Figure 1), together with **SAPA** (**S**patiotemporal **A**ctive-**P**robing **A**gent; Figure 4b), a unified active-perception agent framework. SpaceAct couples closed-loop physical Action with Active perception through four core mechanisms (overview in Figure 2):

1. **Zero-privilege ruler-based localization with an active camera.** No numerical object coordinates are ever given in text. Scenes are instrumented with standard tri-color rulers, projection walls, and visual display boards, and within a 100-turn budget the agent must call `camera(az, el, zoom, pan)` to actively resolve occlusion.
2. **Deliberation-frozen clock with a same-view 5-frame strobe buffer.** To remove differences in API inference latency across models, the world clock $t$ is strictly frozen while the model deliberates ($\mathrm{d}t = 0$) and advances only with physical actions. In dynamic tasks, SAPA renders five strobe frames $\{t - 4\Delta, \ldots, t\}$ ($\Delta = 0.25$ s) from the current camera view, so that velocity, angular velocity, and period can be estimated directly from vision.
3. **Dynamic memory and batched tool calls.** SAPA maintains a dynamic memory: a sliding context window of the 10 most recent text turns and the 2 most recent images prevents context overflow, while a self-managed `note` persists across up to 100 turns to record ruler readings, hidden physical parameters, and intermediate plans. Each turn may issue a batch of 1–5 tool calls, which reduces model API calls by 23%–60% (median 50%).
4. **First-violation truncation, split-screen rollback diagnostics, and controlled attribution modes.** Each action is executed as a fine sweep in substeps of at most 0.05 stud (one stud is the brick grid unit) or 0.5°, without any motion planner, and structures are checked by a calibrated rigid-body equilibrium solver. When a batch collides or loses stability at substep $u^{\star} < 1$, execution is truncated immediately and a 1280 × 640 split-screen rollback view is returned (left: the collision or overload moment; right: the restored valid state). The framework also provides controlled diagnostic switches (state, plan, no-camera, blind, cap1, and six L7 levers) that target visual grounding ($\mathcal{G}_{\text{vis}}$), physical identification ($\mathcal{I}_{\text{phys}}$), multi-step planning ($\mathcal{P}_{\text{plan}}$), and frame-aware control ($\mathcal{E}_{\text{ctrl}}$).

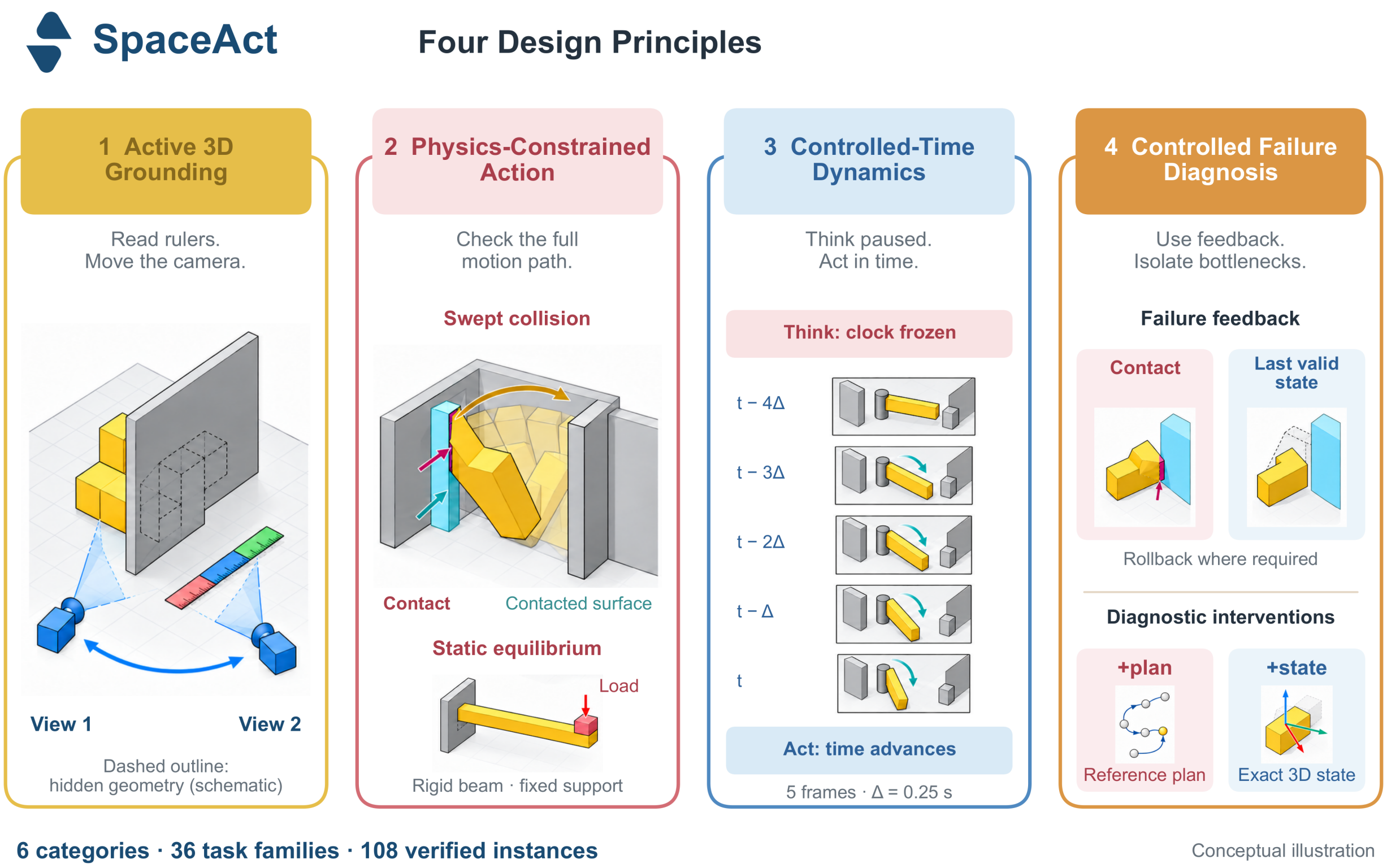

**Figure 2:** SpaceAct's four design principles (conceptual illustration). **Active 3D grounding:** the agent reads in-scene rulers and moves the camera to resolve occlusion, with no privileged coordinates. **Physics-constrained action:** the full swept motion path is checked for collisions and structures for static equilibrium, without a motion planner. **Controlled-time dynamics:** simulation time pauses while the model deliberates and during camera actions, and advances only with timed physical actions; five same-view frames expose the motion. **Controlled failure diagnosis:** split-screen rollback feedback pairs the contact with the last valid state, and separately run +plan and +state conditions isolate the bottleneck.

Built around interlocking bricks and 3D mechanical puzzles, SpaceAct contains 36 task families and 108 procedurally generated instances, three per family at its hardest available tier (L5, or L7 with perception levers), covering six categories (Figure 4a; Section 4).[1] (1) *Structural construction and engineering mechanics* (6 families / 18 instances), checked by a rigid-body statics solver calibrated against StableLego (Liu et al., 2024), covers bridging a deep gorge, load-bearing bridge reduction, anchorless tower cranes, connectivity under a construction-overhang limit, gripper assembly with temporary jig bricks, and counterweighting on a floating-baseplate balance. (2) *Combinatorial planning and puzzles* (7 / 21) requires shortest-path solutions to six-direction occluded disassembly, Klotski and its one-way-vehicle variant, 24-orientation 3D gravity mazes, sliding-bar locks, pattern cubes with occluded bottom faces, and blueprint-specified burr puzzles (Luban locks). (3) *Continuous 6-DoF manipulation* (6 / 18) tests swept-volume control in multi-wall chiral peg-in-hole, corridor and chimney turnarounds, polyline wire loops, twisted double rings, interlocking-heart disentanglement, and borrowed-clearance ring release. (4) *Assembly and packing* (3 / 9) comprises elevation-limited perspective tangrams and opaque deep-box polycube packing. (5) *Multi-view perception and 3D reconstruction* (3 / 9) requires oblique-light shadow-wall construction, low-elevation occluded block counting, and chiral mirror/rotation replication. (6) *Temporal, physical, and mechanism reasoning* (11 / 33) combines 7 strobe-based dynamics and physics-probing families with 4 families of 3D functional mechanism synthesis (bevel-gear transmission, dual-ball marble runs, equal-path-length lasers, and overflow pipe networks). Every instance is validated by a deterministic reference solver and filtered for known shortcuts.

We systematically evaluate 10 frontier MLLMs across SpaceAct and obtain three core findings:

[1] LEGO® is a trademark of the LEGO Group, which does not sponsor, authorize, or endorse this work.

1. **Pronounced capability stratification, with continuous 3D geometric manipulation lagging far behind.** Across the 44 hardest instances, frontier MLLM scores span 4.8–79.1 (success rates 6.8%–93.2%), separating a clear top tier of three models from the middle of the field. Frontier models perform relatively well on dynamics and mechanisms (mean success rate 58.5%) and discrete puzzles (47.1%), but 7 of 10 solve at most 27.3% of continuous 6-DoF manipulation tasks, and 7 solve no multi-view perception task. Adding L7 perception and coordinate-frame levers lowers the mean spatial-manipulation score by 48% (0 of 42 L7 episodes solved outside the top three), whereas undisclosed physical changes in temporal tasks cause no score drop (Sections 6.1–6.3).
2. **Controlled diagnostics show that, even with a correct high-level plan, 3D visual grounding and closed-loop control remain the main bottleneck.** For the three strongest frontier models (22 paired episodes), a reference high-level plan (+plan, bypassing $\mathcal{P}_{\text{plan}}$) raises the score from 65.1 to 80.3, closing 44% of the remaining gap, but does not reduce collisions (5.2 → 5.2). Adding exact 3D state every turn on top of the plan (+state, bypassing $\mathcal{P}_{\text{plan}}$ and $\mathcal{G}_{\text{vis}}$) raises it to 96.6, closing 90% (18 pairs up, 0 down; success rate 100.0%), shortens episodes about sevenfold, and cuts collisions to 0.5; it closes 83% of the gap that the plan alone leaves. The same ordering holds on four further models (ranked 5th–8th): the plan lifts their mean score from 10.9 to 35.3 and increases collisions (21.5 → 28.0 per episode), whereas the plan plus 3D state lifts it to 81.8 (SR 18.8% → 90.6%). Removing images (blind) increases frontier collisions ninefold, and locking the camera (no-camera) lowers the score on occlusion-heavy tasks from 58.3 to 37.8 (Section 6.4).
3. **Observation and collision loops are the most common trajectory-level failure, and rollback feedback rarely translates into geometric correction.** Of the 227 unsolved frontier episodes under the batched protocol, 46% are trapped in collision loops (24%) or observation loops (22%), rising to 60% in 6-DoF manipulation; another 30% stall without a loop, the dominant pattern in 3D mechanism synthesis. Moreover, 44%–71% of frontier collisions are followed by another collision on the very next turn: models grasp the high-level strategy and see the split-screen rollback view, yet struggle to turn contact feedback into accurate metric corrections (Sections 6.5–6.6).

Our main contributions are as follows:

- **SpaceAct, an interactive benchmark for 3D spatial and temporal reasoning.** It spans 6 categories, 36 task families, and 108 graded instances, each verified by a deterministic solver and filtered for shortcuts, and unifies zero-privilege ruler-based localization, motion-planner-free continuous 6-DoF swept manipulation, StableLego-calibrated rigid-body statics, 3D functional mechanism synthesis, and deliberation-frozen 5-frame strobe dynamics in a single closed-loop evaluation.
- **SAPA, an active-perception agent framework.** SAPA (Spatiotemporal Active-Probing Agent) is a model-agnostic agent that perceives and acts through tool calls (active camera control and physical probing alongside manipulation), maintains a dynamic memory (a sliding context window plus a persistent note), observes dynamics through a same-view 5-frame strobe buffer, and self-corrects from magenta/blue split-screen rollback feedback; built-in controlled attribution modes target each operator in the perception–control loop.
- **A systematic evaluation, causal attribution, and trajectory analysis of 10 frontier MLLMs.** We characterize degradation across categories and L7 difficulty levers, and show through nested plan and state injection that, once a correct plan is given, 3D visual grounding and closed-loop control account for most of the remaining gap.

## 2 Related Work

**3D spatial reasoning benchmarks.** Evaluation of spatial abilities in vision-language models relies mainly on static visual question answering. 3DSRBench (Ma et al., 2025a), SpatialRGPT-Bench (Cheng et al., 2024), OmniSpatial (Jia et al., 2026), BLINK (Fu et al., 2024), SpatialVLM (Chen et al., 2024), and the video-based

VSI-Bench (Yang et al., 2025a) probe relative depth, orientation, perspective taking, and metric estimation. Extending toward multi-step reasoning, LEGO-Puzzles (Tang et al., 2026) and SpatialViz-Bench (Wang et al., 2026) test mental rotation and assembly-step prediction with multiple-choice questions. Recent suites broaden coverage to dozens of spatial tasks (SpatialScore; Wu et al., 2026) or to the spatial function of objects (SFI-Bench; Zhang et al., 2026a). In all of these, the model is a passive observer: it never issues continuous 3D actions, never moves the camera to resolve occlusion, and never experiences collision or collapse.

**Table 1:** Comparison with related spatial reasoning and embodied benchmarks (✓ supported, ✗ not supported, ○ partially supported).

| Benchmark scope | | | | | Capabilities | | | | |
|---|---|---|---|---|---|---|---|---|---|
| **Benchmark** | **Category** | **Action level** | **#Fam.** | **#Tasks** | **Closed-loop** | **Zero-priv. 3D** | **3D topology & statics** | **Temporal strobe** | **Control diag.** |
| 3DSRBench (Ma et al., 2025a) | 3D spatial VQA | Passive QA | 12 | 2,772 | × | × | × | × | × |
| VSI-Bench (Yang et al., 2025a) | Video spatial QA | Passive QA | 8 | 5,060 | × | × | × | × | × |
| OmniSpatial (Jia et al., 2026) | Spatial VQA | Passive QA | 50 | 1,533 | × | × | × | × | ✓ |
| LEGO-Puzzles (Tang et al., 2026) | Assembly VQA | Passive QA | 11 | 1,100 | × | × | × | × | ✓ |
| ALFWorld (Shridhar et al., 2021) | Household | High | 6 | 274 | ✓ | × | × | × | × |
| VisualAgentBench (Liu et al., 2025) | Multi-domain | High | 5 | 746 | ✓ | × | × | × | × |
| BALROG (Paglieri et al., 2025) | 2D games | High & low | 6 | 100+ | ✓ | × | × | × | ✓ |
| iVISPAR (Mayer et al., 2025) | 3D sliding puzzle | Low (grid) | 1 | 300 | ✓ | ✓ | × | × | ✓ |
| PhyBlock (Ma et al., 2025b) | Block assembly | Low (grid) | 4 | 400 (+2.2K QA) | ✓ | × | × | × | ✓ |
| BuildArena (Xia et al., 2026) | Bridge building | Low | 3 | 192 | ✓ | × | ✓ | × | × |
| EmbodiedBench (Yang et al., 2025b) | Household & manip. | High & low | 4 | 1,128 | ✓ | × | × | × | ✓ |
| CubeBench (Gao et al., 2026a) | Rubik's Cube | Low (discrete) | 3 tiers | Generative | ✓ | ○ | × | × | ✓ |
| SpatialAct (Liu et al., 2026) | 3D layout repair | High (semantic) | 3 levels | 4,355 | ✓ | ○ | ○ | × | ✓ |
| SpatialWorld (Gao et al., 2026b) | 8 simulators | High (text) | — | 760 | ✓ | ✓ | × | × | × |
| VABench (Zhang et al., 2026b) | Tabletop robot | Low | 14 | 280 | ✓ | ✓ | × | × | ✓ |
| **SpaceAct (ours)** | **Physical & spatiotemporal (6 categories)** | **High & low** | **36** | **108 (procedural)** | ✓ | ✓ | ✓ | ✓ | ✓ |

✓ Yes × No ○ Partial Partial: selected tiers or discrete collision checks.

**Interactive spatial reasoning benchmarks.** Four concurrent works evaluate spatial reasoning in closed loop. CubeBench (Gao et al., 2026a) uses the Rubik's Cube to separate state tracking, spatial reasoning, and exploration under partial observability, and isolates bottlenecks by giving agents solver tools. SpatialAct (Liu et al., 2026) has VLM agents repair 3D layouts over multiple turns through semantic edits (move, rotate, scale) and reports a reasoning-to-action gap relative to humans. SpatialWorld (Gao et al., 2026b) unifies eight simulators under purely visual, partially observable conditions with textual actions. VABench (Zhang et al., 2026b) has MLLMs select camera viewpoints and issue incremental metric end-effector commands for 14 tabletop manipulation families, without privileged object poses, and revise them from execution feedback;

inverse kinematics, collision checking, and trajectory generation are handled by a fixed execution layer. The first three use discrete or semantic actions, abstracting away continuous control, contact, and time. SpaceAct instead requires ruler-grounded metric 6-DoF control, enforces swept collision and statics, introduces a moving world, and separates high-level planning ($\mathcal{P}_{\text{plan}}$) from 3D visual grounding ($\mathcal{G}_{\text{vis}}$) by injecting a reference plan, and then exact 3D state on top of it, into matched episodes. Our findings are consistent with the reasoning-to-action gap reported by SpatialAct and, in continuous control, locate most of the gap that remains after planning in metric grounding and closed-loop control.

**Interactive embodied and puzzle benchmarks.** ALFWorld (Shridhar et al., 2021), VisualAgentBench (Liu et al., 2025), and EmbodiedBench (Yang et al., 2025b) place MLLMs in household and tabletop simulators, while BALROG (Paglieri et al., 2025) and iVISPAR (Mayer et al., 2025) test multi-step planning on games and sliding puzzles. These benchmarks sidestep the hard parts of 3D physics in three ways. First, semantic priors dominate high-level household tasks, so removing images barely affects performance. Second, in low-level manipulation, object coordinates come from external detectors or visual marker prompting, and a built-in motion planner computes collision-free trajectories. Third, time has no effect on their tasks: nothing moves on its own, so no multi-frame temporal estimation is needed. SpaceAct removes all three shortcuts.

**Physical construction and intuitive physics.** PHYRE (Bakhtin et al., 2019), I-PHYRE (Li et al., 2024), DeepPHY (Xu et al., 2026), Physion (Bear et al., 2021), and CLEVRER (Yi et al., 2020) focus on passive outcome prediction or 2D trial and error. Recent block-building benchmarks, including PhyBlock (Ma et al., 2025b) and BuildArena (Xia et al., 2026), introduce stepwise assembly and load testing. SpaceAct brings together, in a single benchmark, zero-privilege ruler-based localization, continuous topological and volumetric manipulation, StableLego-calibrated statics and 3D functional mechanisms, deliberation-frozen strobe dynamics, and controlled diagnostic modes (Table 1).

**Feedback-driven LLM agents.** SAPA builds on reasoning–acting agents (ReAct; Yao et al., 2023) and self-correction from environment feedback (Reflexion, Shinn et al., 2023; Inner Monologue, Huang et al., 2022). Unlike general-purpose frameworks, it unifies first-violation truncation and split-screen rollback views, the two most critical forms of visual feedback in physical tasks, and thereby evaluates every model under identical information conditions.

## 3 Problem Formulation and the SAPA Framework

### 3.1 Action Space

Actions in embodied systems span multiple levels of abstraction. Rather than restricting the agent to high-level semantic macros or predefined keypoints, SpaceAct defines a three-tier action space $\mathcal{A} = \mathcal{A}_{\text{sense}} \cup \mathcal{A}_{\text{kin}} \cup \mathcal{A}_{\text{sys}}$.

- **Active perception and probing ($\mathcal{A}_{\text{sense}}$):** information-gathering actions that do not change the terminal state. A camera action $a^{\text{cam}} = (\text{az}, \text{el}, \text{zoom}, p)$ updates the projection $\Pi(c)$ while world time stays unchanged ($\Delta t = 0$). Probing actions (`probe`, `test_load`, `crank`, `topple`) excite the environment to reveal hidden parameters, for example by rolling a test ball through a hidden magnetic field or cranking a gear train to check its output.
- **Metric kinematic and construction primitives ($\mathcal{A}_{\text{kin}}$):** rigid-body translation and rotation, grasp switching, or brick placement, specified in the world frame $\mathcal{F}_W$ or the payload's own frame $\mathcal{F}_T$: $a^{\text{kin}} = (\Delta x, \Delta y, \Delta z, \hat{\text{u}}, \Delta\theta, \text{part}, \text{grip})$. The payload is never teleported by a motion planner; instead, it is checked along a continuous path $\gamma(u) \in \text{SE}(3)$ in fine steps ($\delta d \leq 0.05$ stud, $\delta\theta \leq 0.5°$) and stops at the first violation $u^\star$ given by Eq. (1), where $\Phi_{\text{free}}$ checks geometric collision and topological constraints (including gripper clearance and rope length) and $\Phi_{\text{static}}$ checks rigid-body equilibrium, the in-construction overhang limit, or the admissible center-of-mass range.

$$u^\star = \sup\{u \in [0,1] \mid \forall v \le u \colon \Phi_{\text{free}}(\gamma(v)) = 1 \wedge \Phi_{\text{static}}(\gamma(v)) = 1\} \tag{1}$$

- **Combinatorial, temporal, and batched actions ($\mathcal{A}_{\text{sys}}$):** discrete puzzle actions (`slide`, `tilt`, `move_disk`), timed actions (`wait` $\Delta t$, `thrust` $(F, \Delta t)$, timed moves), and one-shot commit actions (`release`, `open_valve`, `answer`, `submit`) whose outcome is final. At each turn the agent outputs a batch $\mathcal{B}_\tau$ of length $1 \le k \le 5$, executed in order until all actions succeed or the first violation occurs at action $j^\star$, after which the remaining actions are skipped.

### 3.2 A Deliberation-Frozen Spatiotemporal POMDP

We model vision-driven physical problem solving as a partially observable decision process $\mathcal{M} = (\mathcal{S}, \Theta, \mathcal{A}, \mathcal{C}, \Omega, \mathcal{T}_\theta, \mathcal{O}, L, \mathcal{R})$ with an explicit simulation clock $t$. Here $\mathcal{S}$ is the 3D geometric and kinematic state; $\Theta$ collects hidden parameters that cannot be read from a single frame (e.g., occluded internal bricks, the notches of a burr puzzle, joint pull-out strength, ice and snow friction, hidden magnetic fields, and the angular velocity of rotating gates); $\mathcal{C}$ is the camera-pose space; $\Omega$ and $\mathcal{O}$ are the observation space and rendering function; $\mathcal{T}_\theta$ is the transition function under hidden parameters $\theta \in \Theta$; $L$ is the task instruction; and $\mathcal{R}$ is the score. The simulation clock is frozen during model deliberation ($\mathrm{d}t = 0$) and advances only with action execution:

$$t_\tau = t_{\tau-1} + \sum_{i \le j^\star} \mathrm{dur}(a_{\tau,i}) \tag{2}$$

Each observation is either a single frame $I_\tau$ or five strobe frames (interval $\Delta = 0.25$ s) rendered from the current camera pose (full-resolution examples in Figure A.2). Conditioned on the instruction $L$, the persistent note $m_{\tau-1}$, and the bounded history $h_\tau$, the policy outputs $(m_\tau, \mathcal{B}_\tau)$ so as to maximize the expected score $\mathcal{R}$.

### 3.3 Diagnostic Decomposition

Solving an episode requires chaining four sub-capabilities, which we write conceptually as a composition of operators in the per-turn policy:

$$\pi = \mathcal{E}_{\text{ctrl}} \circ \mathcal{P}_{\text{plan}} \circ \left(\mathcal{G}_{\text{vis}}, \mathcal{I}_{\text{phys}}\right) \tag{3}$$

Eq. (3) organizes the controlled diagnostics, whose inputs are summarized in Figure 3: each intervention replaces one sub-capability with its ground truth or subjects it to a controlled constraint.

- $\mathcal{G}_{\text{vis}}$ **3D grounding:** inverting the perspective projection against the rulers and fusing multiple views to recover coordinates and occluded structure. The +state mode supplies the true state on top of the reference plan; the no-camera mode locks the camera at its initial pose, the blind mode removes images, and the +occ and +blueprint levers make the image-to-coordinate mapping harder.
- $\mathcal{I}_{\text{phys}}$ **physical identification:** estimating velocities and periods from strobe frames, or inferring hidden parameters from probes and rollbacks; no mode replaces it, and the undisclosed physical changes of the D3 and D5 L7 tiers test it.
- $\mathcal{P}_{\text{plan}}$ **multi-step planning:** searching the configuration space under disentanglement, chirality, temporary-support, and balance constraints; the plan mode provides a reference plan.
- $\mathcal{E}_{\text{ctrl}}$ **frame-aware control:** mapping the plan to action vectors in rotated or payload frames and self-correcting from rollback views; stressed by the +tool, +rot, and +nonum levers and the cap1 mode.

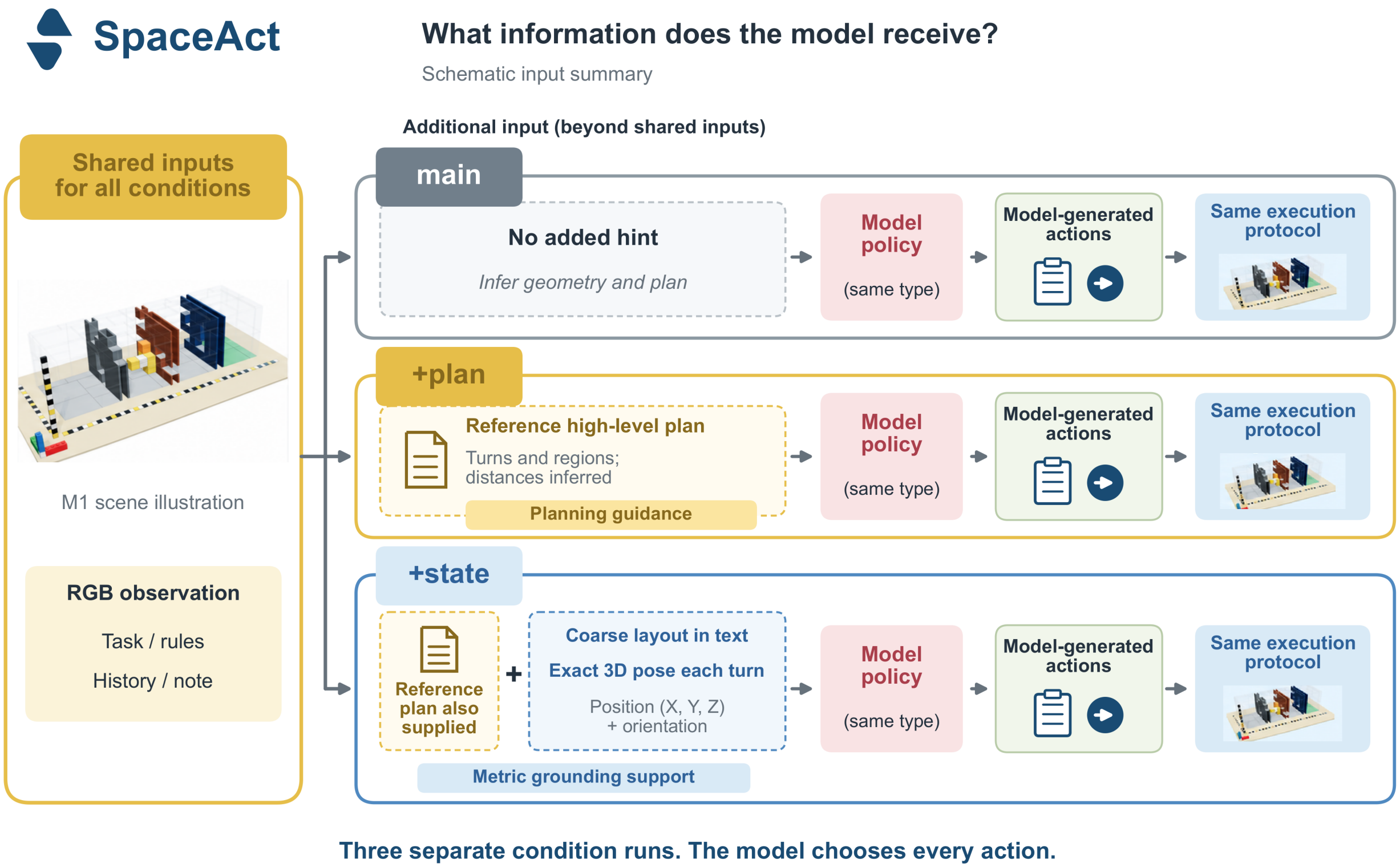

**Figure 3:** Inputs of the three controlled-diagnostic conditions (schematic), each run separately on the same instances. All conditions share the RGB observation, task and rules, history, and note; the model generates every action, and all actions run under the same execution protocol. +plan adds a reference high-level plan (turns and regions; distances are left to the model). +state supplies the reference plan as well, plus the coarse layout as text and the exact 3D pose of the payload at every turn.

### 3.4 SAPA: An Active-Perception Agent

SAPA is an active-perception agent: at every turn it decides where to look and what to do through tool calls, and it carries a dynamic memory across turns. Figure 4b–c illustrates one SAPA turn, which consists of three stages:

- **Observation and dynamic memory.** At each turn $\tau \in \{1, \ldots, 100\}$, the agent receives (i) a visual observation, namely a rendered image with rulers and projection walls or a 5-frame strobe strip; (ii) the task goal, physical rules, JSON response format, and the turn counter `=== Turn τ / 100 ===`; (iii) the tools available for the task (its action primitives), plus any enabled diagnostic mode or lever; and (iv) its dynamic memory: a sliding window of the last 10 text turns and the last 2 images, together with the persistent note $m_{\tau-1}$ that the agent rewrites itself.
- **Reasoning and tool calls.** The model replies with a single JSON object `{"reason", "note", "actions"}`: a free-text rationale, a memory update (the new note, which overwrites the previous one), and 1–5 structured tool calls. Perception tools (camera moves and probes) gather information, and manipulation tools (translations, rotations, grasp switches, brick placements, and timed actions) change the scene. The framework prescribes no reasoning steps; grounding, temporal estimation, reflection on the last rollback, and planning all take place within this single call.
- **Execution and tool feedback.** The environment executes the tool calls in order. If a call violates $\Phi_{\text{free}}$ or $\Phi_{\text{static}}$ at $u^{\star} < 1$, execution stops, the remaining calls are marked as skipped, the state is rolled back if necessary, and the next frame is a 1280 × 640 split screen: the left half shows the violation moment (the offending part in magenta, the obstacle or overloaded joint in bright blue), and the right half shows the valid state after rollback. Textual feedback reports the result of every call.

### 3.5 Scoring

Each episode receives a tiered completion-quality score $S \in [0, 100]$:

$$S = \begin{cases} 40 + 60\sqrt{\min\left(1, \dfrac{A^\star}{A_{\text{used}}}\right) \cdot 0.88^{C} \cdot \left(\min\left(1, \dfrac{\max(3, A^\star)}{T_{\text{used}}}\right)\right)^{1/4}}, & \text{if solved,} \\ 35 \cdot P_{\max} \cdot 0.92^{C}, & \text{otherwise,} \end{cases} \tag{4}$$

where $A^\star$ is the number of actions in the reference solution, $A_{\text{used}}$ is the number of scored actions (camera actions excluded), $C$ is the number of collisions or rollbacks, $T_{\text{used}}$ is the number of turns used, and $P_{\max}$ is the maximum progress achieved. Solved episodes score in [40, 100] and unsolved episodes in [0, 35]. Construction tasks that end in a load test (B1, B2, B3) are graded pass/fail and contribute only to the success rate.

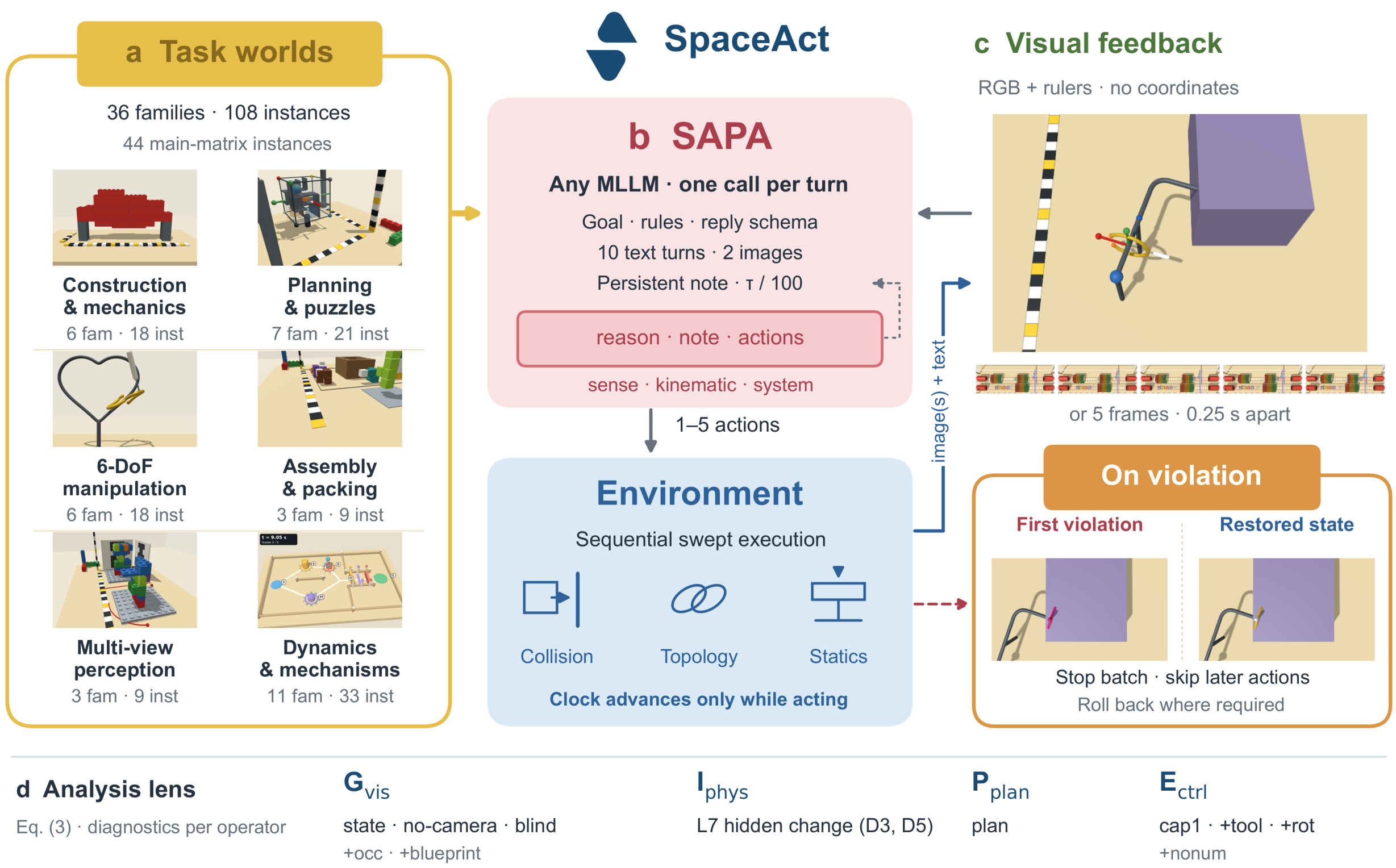

**Figure 4:** The SpaceAct task suite and the SAPA active-perception agent. **(a)** The six task categories contain 36 task families and 108 verified instances in total (44 in the main matrix). **(b)** At each turn, the input comprises actively acquired RGB observations with rulers, task rules, the available tools, and a dynamic memory (a bounded history window plus a persistent note). A single MLLM call per turn jointly performs 3D localization ($\mathcal{G}_{\text{vis}}$), physics identification ($\mathcal{I}_{\text{phys}}$), multi-step planning ($\mathcal{P}_{\text{plan}}$), and frame-aware control ($\mathcal{E}_{\text{ctrl}}$), returning a JSON response with a rationale, a memory update (note), and 1–5 tool calls; the four capabilities are a diagnostic decomposition, not four separate calls. The tool calls execute sequentially in swept substeps of at most 0.05 stud or 0.5°, without a motion planner, under collision/topology and static-equilibrium checks, and the simulation clock is frozen while the model deliberates. **(c)** Observations are RGB images with rulers and no coordinates; dynamic tasks provide five frames from the same viewpoint at 0.25-s intervals. The first violation truncates execution and skips the remaining calls, returning a 1280 × 640 split-screen rollback view (left: the violating contact; right: the last valid state). **(d)** The diagnostic conditions grouped by the operator of Eq. (3) that they replace or stress.

## 4 The SpaceAct Benchmark

### 4.1 Six Task Categories

SpaceAct contains 36 task families and 108 instances (three per family), grouped into six categories (Figure 4a and Table 2; full descriptions in Appendix A).

- **Structural construction and engineering mechanics (6 families / 18 instances).** Agents build or reduce brick structures under checks by a StableLego-calibrated rigid-body equilibrium (RBE) solver, which agrees with a Gurobi reference (Gurobi Optimization, LLC, 2025) on 99.85% of cases. Tasks include spanning a deep gorge between slender 2×2 piers under a three-zone moving load (B1); reducing a bridge to a minimal skeleton without collapse (B2); extending the jib of an anchorless tower crane without tipping at any step (B3); connecting upright and inverted anchors under a 3-stud construction-overhang limit (B4); inserting and later extracting temporary jig bricks during two-finger-gripper assembly (B5); and replicating a cantilevered sculpture on a baseplate floating on a central pivot, using counterweights in stages (B6).
- **Combinatorial planning and puzzles (7 / 21).** Discrete puzzles are judged against the shortest solution: six-direction occluded disassembly with decoys and a deeply buried target (P1); Klotski traffic escape (P2) and its one-way-vehicle variant (P3); an opaque 3D gravity maze with 24 orientations (P4); a three-axis interleaved sliding-bar lock (P5); a 3×3×3 pattern cube whose bottom face can only be inferred from the parts list (P6); and a burr puzzle whose notches appear only on an in-scene blueprint and whose hardest tier requires a push-then-retract move (P7).
- **Continuous 6-DoF manipulation (6 / 18).** A gripper holds a rigid body and, without any motion planner, passes it through near-zero-clearance obstacles via continuous translations and rotations: a chiral polycube through up to three silhouette walls with decoy holes (M1); an eccentric long rod turned around in a glass corridor via a turning chamber or a vertical chimney (M2); a C-ring guided along a polyline wire around supports (M3); a double ring rolled along a twisted wire (M4); two interlocking hearts separated through misaligned gaps (M5); and an open ring extracted after switching grasp to move an auxiliary ring aside (M6).
- **Assembly and packing (3 / 9).** Under oblique views (elevation ≤ 60°), agents pick, rotate, flip, and place tangram pieces into a recess drawn only as an outline, with surplus pieces, rotated frames, and targets given only on a display board (A1, A2); they also rotate polycubes about three axes and pack them into an opaque box whose lower layers are occluded, while rejecting mirror-image decoys (A3).
- **Multi-view perception and 3D reconstruction (3 / 9).** No single view suffices: agents build a structure whose shadows on the north wall, the east wall, and under oblique light all match target silhouettes, which requires overhangs to open window holes (V1); count the bricks of an irregular pile while orbiting at elevation ≤ 40°, inferring hidden columns from support rules (V2); and replicate a mirrored or rotated version of a multicolor model whose buried core is revealed by back-wall silhouettes (V3).
- **Temporal, physical, and mechanism reasoning (11 / 33).** Seven families depend on strobe observation or physical probing: helical key extraction (D1), a three-gate phase lock (D2), airdrops into a tunnel with segment-wise speeds (D3), dual-track duck avoidance with puddles (D4), braking on ice around a tower with undisclosed snow and tilt (D5), spatiotemporal reversal on a 7×7 cloud track (D6), and an occluded helical wire visible only through a rotated mirror (D7). Four further families require reading targets from in-scene boards and scales and building a working 3D mechanism: a gear train with bevel-gear reversal and phase gates (D8), a gravity marble run with pedal gates and a bistable switch (D9), equal-path-length lasers via mirrors and prisms (D10), and pipe-network water distribution with splitting and overflow (D11).

**Table 2:** Composition of the six SpaceAct task categories. *Main* is the number of instances in the main evaluation matrix.

| Category | Action level | Families | #Fam. | #Inst. | Main |
|---|---|---|---|---|---|
| Structural construction & engineering mechanics | Low & high | B1-B6 | 6 | 18 | 6 |
| Combinatorial planning & puzzles | High & low | P1-P7 | 7 | 21 | 7 |
| Continuous 6-DoF manipulation | Low (6-DoF) | M1-M6 | 6 | 18 | 11 |
| Assembly & packing | Low (voxel) | A1, A2, A3 | 3 | 9 | 4 |
| Multi-view perception & 3D reconstruction | Low & high | V1-V3 | 3 | 9 | 3 |
| Temporal, physical & mechanism reasoning | High & low | D1-D11 | 11 | 33 | 13 |
| **Total** | | | **36** | **108** | **44** |

Main = instances in the main evaluation matrix (44 total). Family ranges are inclusive.

### 4.2 Difficulty Tiers and L7 Levers

Parameterized families are graded along family-specific difficulty axes. L1 has one or two constraints with generous geometric or temporal margins; L3 couples several constraints and adds decoys or one non-monotone dependency; L5 tightens tolerances and adds chiral decoys, oblique projections, or multi-branch synchronization. L7 keeps the L5 task fixed and adds perception and coordinate-frame levers: +nonum (blocked actions report only "stopped early"), +occ (opaque occluding panels), +tool (actions specified along the payload's own axes), +rot (the track or target board is rotated and offset from the grid), +blueprint (part of the track is hidden in a tunnel and only a half-scale blueprint is provided), and, for the tangram family A1, +poster (the target is shown only on a back-wall poster). For the temporal families D3 and D5, L7 instead introduces undisclosed physical changes, such as unreported tunnel speeds or a slight tilt of the ice. The released library contains the hardest available tier of every family.

### 4.3 Capability Annotation

Each family is annotated with fine-grained capability tags, aggregated into six cross-category capabilities: metric and multi-view 3D grounding; mental rotation and chirality discrimination; topological and swept-volume constraints; non-monotone sequencing (temporary jigs, counterweights, push-then-retract moves, one-way rules); quantitative statics; and temporal estimation with mechanism synthesis.

### 4.4 Verification

Every instance is verified to be solvable by program, and its reference solution defines the action count $A^{\star}$ used in scoring. For discrete puzzles, search algorithms (e.g., BFS, Dijkstra, or exact cover) compute an optimal solution and, where the task requires it, certify that the answer is unique; continuous manipulation and construction tasks are verified by replaying scripted reference solutions or by their load checkers. Generators also reject instances that admit known shortcuts. Solver and filtering details are given in Appendix A.

## 5 Experimental Setup

**Models.** We evaluate 10 frontier MLLMs through public APIs: Claude Opus 5.5, GPT-6 Astra, Gemini 4 Argon, Muse Wonder 1.0, Doubao Seed 2.1 Pro, Qwen 3.8 Max, DeepSeek V4.1 Flash, Kimi K3, MiMo V2.6

Pro, and Step 5 Preview. Each model uses the reasoning setting listed in Appendix B, with default sampling parameters.

**Main matrix.** The main matrix targets the hardest setting of each family: the hardest tier of every parameterized family (L5, plus L7 for the eight families that have it), together with the single instance of each fixed family and A2, which has only an L3 tier. This yields 44 instances covering all 36 families (Build 6, Puzzle 7, Manipulation 11, Assembly 4, Perception 3, Dynamics 13). Each model runs once per instance (seed 100) with a 100-turn budget and up to 5 actions per turn.[2]

**Metrics.** The primary metric is the tiered score (Eq. 4) averaged over the 41 instances with graded scores. We also report the success rate (SR) and, per episode, the average number of turns, collisions/rollbacks, and total tokens. At the category level we report within-category SR.

**Baselines.** A random valid agent samples uniformly among valid actions; a scripted myopic greedy agent maximizes immediate progress at each step (covering 29 of the 44 instances); and the reference solver replays the verified optimal solution (covering 38 of the 44 instances, with the BrickAgent-based tasks verified by their own checkers).

**Controlled diagnostics.** Diagnostics are run on the three strongest models (Claude Opus 5.5, GPT-6 Astra, and Gemini 4 Argon), all on L5 instances, with each mode targeting one sub-capability in Eq. (3); plan and state also cover four further models ranked 5th–8th in Table 3 (Doubao Seed 2.1 Pro, Qwen 3.8 Max, DeepSeek V4.1 Flash, and Kimi K3), and Appendix C lists the other 7-model ablations. The instance sets are plan and state on M1, M2, M3, M4, A3, P4, D8, and A1; blind on B1, M1, M3, B6, and P2; no-camera on A3, V1, P6, V2, V3, B5, P7, and M2; and cap1 on M1, M3, P4, A1, D2, D3, and D5. We further include 16 single-factor lever–instance combinations and additional runs with seeds 101 and 102 on six families. All comparisons are paired against the main-matrix episode of the same model on the same instance. For every model, an episode that ended after six consecutive failed API calls (errors, timeouts, or empty responses) was discarded and restarted from scratch, and only the final complete run is scored; Claude Opus 5.5 needed restarts on 17 of its 44 main-matrix cells (14 then completed) and every other model on at most 6. Twenty Claude Opus 5.5 runs (three in the main matrix, fifteen in diagnostic conditions, and two in the seed reruns) could not be completed within 3–12 attempts because the endpoint returned empty streamed responses, connection errors, or server errors (median 9.9 minutes per failed call); no call reached the output-token limit. Most of these failures fell in one 12-hour period in which only 4 of 101 Claude attempts completed across all conditions (0 of 17 in the main matrix); all of Claude's nonote and nosplit runs were scheduled in that period. In the main matrix the abandoned runs count as unsolved with score 0 (Table 3). In the paired diagnostics we drop every pair in which either run was abandoned (2 of 24 top-3 pairs in each of +plan and +state, 5 of 24 no-camera pairs, 4 of 42 pairs in each of nonote and nosplit, and 2 of 21 cap1 pairs), so that no condition is credited or blamed for an endpoint failure.

## 6 Results

### 6.1 Main Results

Table 3 reports the main results. SpaceAct separates models sharply: across the ten frontier models, tiered scores range from 4.8 (Step 5 Preview) to 79.1 (Claude Opus 5.5), with a median of 23.1, and fall into three clear tiers. The top three models form the leading tier: Claude Opus 5.5 averages 79.1 (SR 93.2%), GPT-6 Astra 65.0 (84.1%), and Gemini 4 Argon 42.1 (65.9%), whereas the remaining seven models average 16.6 (SR 25.6%). Models ranked 4th–7th score 19.5–29.0, and the remaining three score 4.8–10.4. Every frontier model scores above the random valid baseline (score 3.2, SR 6.8%), although the two weakest models only match its success rate, and the reference solver solves every instance it covers. No main-matrix instance is solved by all ten frontier models, and the opaque 4×4×4 3D gravity maze (P4-L5) is solved by none of the nine models whose run completed (Claude Opus 5.5's run was abandoned; Section 5).

[2] The six construction and puzzle tasks built on the BrickAgent environment (Kulits et al., 2026), namely B1, B2, B3, P1, P2, and P3, use a single-step version of the protocol with the same observation and feedback design.

We adopt the tiered score as the primary metric because it resolves finer differences than SR: surplus actions, collisions, and turns are all penalized in solved episodes, so even Claude Opus 5.5 reaches 79.1, leaving ample headroom at the top of the benchmark. The strongest models also solve faster (Figure 5b): Claude Opus 5.5 solves 86.4% of instances within 30 turns and GPT-6 Astra 59.1%, while most other models keep solving instances until close to the 100-turn cap. Solution quality is not simply bought with tokens: Claude Opus 5.5 generates about 26.3K completion tokens per turn but needs very few turns, whereas GPT-6 Astra has the lowest total token usage per episode of all models.

**Table 3:** Main results on the 44-instance main matrix (one run per model and instance, seed 100). Score is the tiered score averaged over the 41 graded instances; SR is the success rate over all 44 instances; category columns give within-category SR (%), with instance counts in parentheses; Turns, Coll., and Tokens are per-episode means. Baselines cover the indicated subsets. Bold marks the best frontier-model score, SR, and category SR (ties included).

| | Overall | | Category success rate (%) | | | | | | Episode cost | | |
|---|---|---|---|---|---|---|---|---|---|---|---|
| Model | Score | SR (%) | Build (6) | Puzzle (7) | Manip. (11) | Assem. (4) | Percep. (3) | Dyn. (13) | Turns | Coll. | Token (M) |
| **Claude Opus 5.5** | **79.1** | **93.2** | **100.0** | **85.7** | **90.9** | 75.0 | **100.0** | **100.0** | 16.1 | 1.8 | 0.67 |
| GPT-6 Astra | 65.0 | 84.1 | 83.3 | **85.7** | 63.6 | **100.0** | **100.0** | 92.3 | 37.1 | 4.9 | 0.29 |
| Gemini 4 Argon | 42.1 | 65.9 | 50.0 | 57.1 | 63.6 | 75.0 | 33.3 | 84.6 | 49.4 | 12.1 | 1.43 |
| Muse Wonder 1.0 | 29.0 | 43.2 | 16.7 | 71.4 | 18.2 | 0.0 | 0.0 | 84.6 | 72.9 | 14.7 | 1.12 |
| Doubao Seed 2.1 Pro | 24.7 | 38.6 | 33.3 | 57.1 | 18.2 | 25.0 | 0.0 | 61.5 | 70.0 | 16.8 | 1.04 |
| Qwen 3.8 Max | 21.5 | 36.4 | 16.7 | 28.6 | 27.3 | 50.0 | 0.0 | 61.5 | 63.8 | 21.4 | 1.18 |
| DeepSeek V4.1 Flash | 19.5 | 31.8 | 16.7 | 57.1 | 9.1 | 25.0 | 0.0 | 53.8 | 74.7 | 17.9 | 2.31 |
| Kimi K3 | 10.4 | 15.9 | 16.7 | 28.6 | 0.0 | 25.0 | 0.0 | 23.1 | 74.0 | 27.7 | 1.00 |
| MiMo V2.6 Pro | 6.1 | 6.8 | 16.7 | 0.0 | 0.0 | 0.0 | 0.0 | 15.4 | 79.1 | 25.1 | 1.37 |
| Step 5 Preview | 4.8 | 6.8 | 33.3 | 0.0 | 0.0 | 0.0 | 0.0 | 7.7 | 85.9 | 23.8 | 2.45 |
| **Baselines and reference solver** | | | | | | | | | | | |
| Random valid | 3.2 | 6.8 | 16.7 | 14.3 | 0.0 | 0.0 | 0.0 | 7.7 | 81.0 | — | — |
| Myopic greedy (29/44) | 4.1 | 6.9 | 0.0 | 0.0 | 18.2 | 0.0 | 0.0 | 0.0 | 87.8 | — | — |
| Reference solver (38/44) | 100.0 | 100.0 | 100.0 | 100.0 | 100.0 | 100.0 | 100.0 | 100.0 | 3.4 | — | — |

Category columns report within-category SR (%). Bold: best evaluated-model score/SR, including ties.

Baseline coverage is retained in parentheses; unreported collisions and tokens are shown as dashes.

**Takeaway.** SpaceAct produces a steep gradient across frontier models (scores 4.8–79.1), with a clear leading tier and a distinct middle tier. Because the tiered score jointly penalizes surplus actions, collisions, and turns, even the strongest model (79.1) leaves substantial room for improvement.

## 6.2 Category Profiles

**(a) SR (%) by category**

| | Build | Puzzle | Manip. | Assem. | Percep. | Dyn. |
|---|---|---|---|---|---|---|
| Claude Opus 5.5 | 100 | 86 | 91 | 75 | 100 | 100 |
| GPT-6 Astra | 83 | 86 | 64 | 100 | 100 | 92 |
| Gemini 4 Argon | 50 | 57 | 64 | 75 | 33 | 85 |
| Muse Wonder 1.0 | 17 | 71 | 18 | 0 | 0 | 85 |
| Doubao Seed 2.1 Pro | 33 | 57 | 18 | 25 | 0 | 62 |
| Qwen 3.8 Max | 17 | 29 | 27 | 50 | 0 | 62 |
| DeepSeek V4.1 Flash | 17 | 57 | 9 | 25 | 0 | 54 |
| Kimi K3 | 17 | 29 | 0 | 25 | 0 | 23 |
| MiMo V2.6 Pro | 17 | 0 | 0 | 0 | 0 | 15 |
| Step 5 Preview | 33 | 0 | 0 | 0 | 0 | 8 |

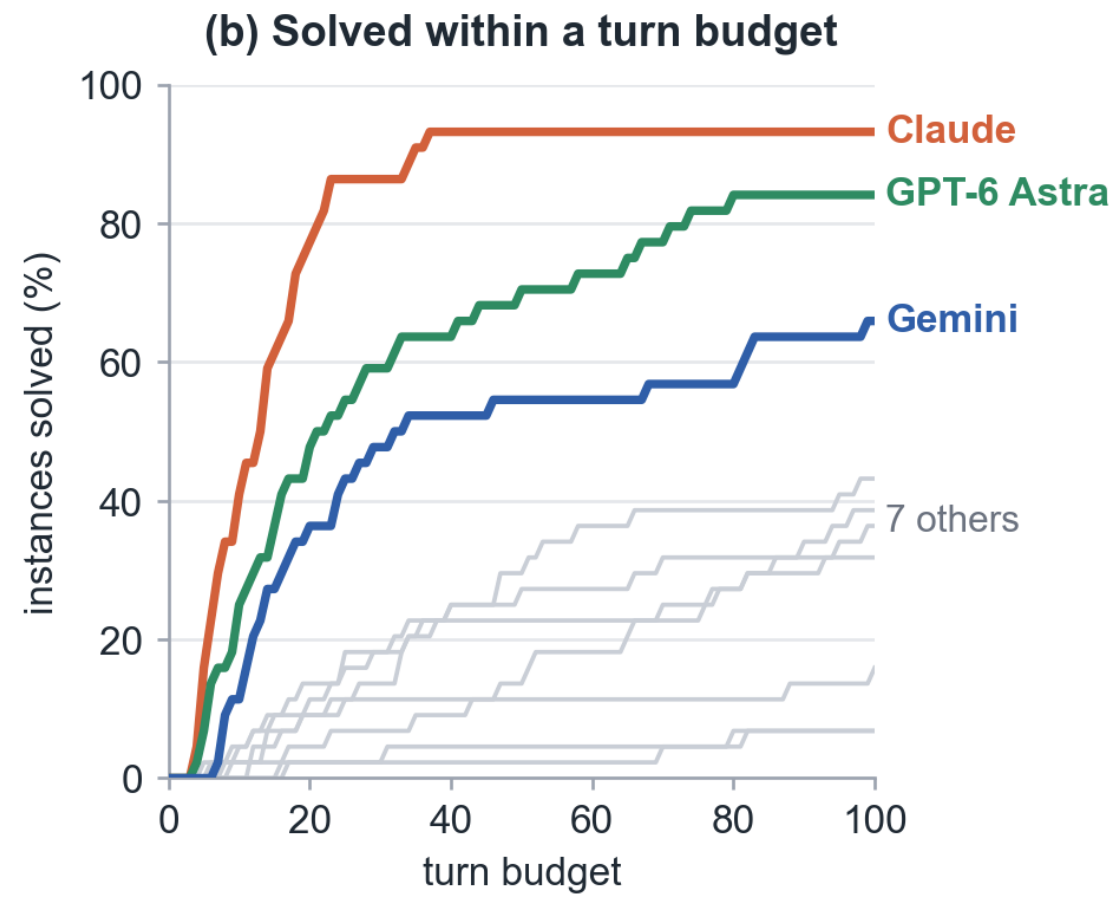

**Figure 5:** Category profiles and solving speed. **(a)** Success rate (%) of each model by task category on the main matrix. **(b)** Cumulative fraction of main-matrix instances solved within a given turn budget; the three strongest models are highlighted and the remaining seven are shown in gray.

Figure 5a profiles success rates by category. Averaged over models, dynamics and mechanism tasks are the easiest (SR 58.5%), followed by puzzles (47.1%), construction (38.3%), assembly (37.5%), 6-DoF manipulation (29.1%), and multi-view perception (23.3%). The two hardest categories are exactly the two that require recovering metric 3D geometry from images: 7 of 10 models solve at most 27.3% of manipulation tasks, and 7 solve no perception task. Category profiles are correlated but not interchangeable: the mean pairwise Kendall's $\tau_a$ between per-category model rankings is 0.53 (minimum 0.38). Claude Opus 5.5 and GPT-6 Astra rank in the top two in every category (Gemini 4 Argon ties for second on manipulation and assembly), while the rest of the field reorders; for example, Muse Wonder 1.0 ranks third on puzzles, ahead of Gemini 4 Argon, and ties for third on dynamics, yet sits near the bottom on construction and assembly. A single aggregate score therefore hides capability differences that the category profiles reveal.

> **Takeaway.** 6-DoF manipulation and multi-view perception, the two categories that require recovering metric 3D geometry from pixels, are the hardest; dynamics and mechanism tasks are comparatively easy.

## 6.3 Difficulty Levers

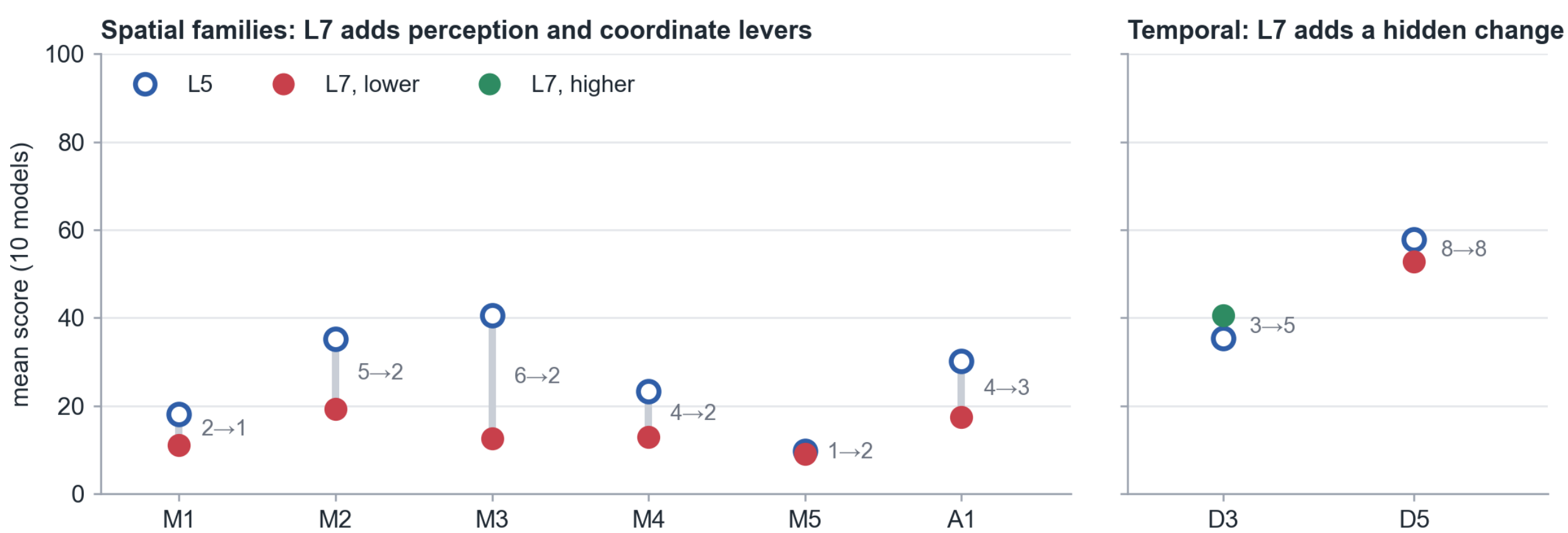

**Figure 6:** Difficulty levers. Mean score over the 10 models at L5 (open circles) and L7 (filled; red if lower, green if higher) for the eight families that include both tiers; annotations give the number of solved episodes at L5 → L7. Left: spatial families, where L7 adds perception and coordinate-frame levers. Right: temporal families, where L7 adds an undisclosed physical change.

Eight main-matrix families include both L5 and L7 (Figure 6). In the six spatial families (M1, M2, M3, M4, M5, A1), where L7 adds perception and coordinate-frame levers, the mean score drops from 26.2 to 13.7 (a relative drop of 48%) and the number of solved episodes falls from 22 to 12; pairing by model and family, L7 lowers the score in 37 of 60 pairs and raises it in 20. The drop is concentrated outside the leading tier: the seven models below the top three solve 0 of 42 L7 spatial episodes, and interlocking-heart disentanglement (M5) is the lowest family at both tiers (mean score below 10). In contrast, the temporal families whose L7 adds undisclosed physical changes (D3, D5) show no drop (mean 46.5 at L5 vs. 46.6 at L7), indicating that models adapt to hidden physical changes after interactively probing the environment. Because L7 leaves the underlying task, and hence its plan, unchanged, the spatial drop isolates the cost of perception and coordinate-frame transformation rather than additional environment dynamics.

**Takeaway.** Hiding numerical feedback, rotating the track or board off the grid, or moving geometry into a blueprint degrades performance far more than changing the environment's physics; because L7 keeps the plan fixed, this cost comes from perception and coordinate-frame transformation.

## 6.4 Controlled Diagnostics

**Table 4:** Controlled diagnostics on L5 instances across the top-3 frontier models (Claude Opus 5.5, GPT-6 Astra, Gemini 4 Argon) and four further models ranked 5th–8th (Doubao Seed 2.1 Pro, Qwen 3.8 Max, DeepSeek V4.1 Flash, Kimi K3). Each entry is a paired comparison, main → diagnostic condition.

| Condition | #Inst. | Pairs n | Score | SR (%) | Pairs ↑ / ↓ | Turns | Coll. |
|---|---|---|---|---|---|---|---|
| **Input ablations** | | | | | | | |
| **No image input ‡**<br>Text feedback only; top 3 | 5 | 15 | 73.9 → 17.7 | 86.7 → 53.3 | 0 / 12 | 45.6 → 59.1 | 3.6 → 33.5 |
| **Camera locked at initial view ‡**<br>Top 3 frontier models | 8 | 19 | 58.3 → 37.8 | 68.4 → 47.4 | 5 / 14 | 38.6 → 43.5 | 5.6 → 5.7 |
| **No persistent note ‡**<br>nonote; all 7 models | 6 | 38 | 31.8 → 29.5 | 47.4 → 42.1 | 11 / 17 | 60.1 → 67.5 | 14.6 → 17.8 |
| **No split rollback view ‡**<br>nosplit; all 7 models | 6 | 38 | 31.8 → 30.3 | 47.4 → 42.1 | 9 / 20 | 60.1 → 64.9 | 14.6 → 23.2 |
| **Top 3 frontier models** | | | | | | | |
| **+plan**<br>Reference high-level plan | 8 | 22 | 65.1 → 80.3 | 81.8 → 95.5 | 10 / 8 | 36.7 → 16.6 | 5.2 → 5.2 |
| **+state**<br>Plan + exact 3D state each turn | 8 | 22 | **65.1 → 96.6** | **81.8 → 100.0** | **18 / 0** | **36.7 → 5.3** | **5.2 → 0.5** |
| **4 further models (ranks 5–8)** | | | | | | | |
| **+plan**<br>Reference high-level plan | 8 | 32 | 10.9 → 35.3 | 18.8 → 43.8 | 19 / 11 | 80.4 → 62.5 | 21.5 → 28.0 |
| **+state**<br>Plan + exact 3D state each turn | 8 | 32 | **10.9 → 81.8** | **18.8 → 90.6** | **30 / 2** | **80.4 → 16.6** | **21.5 → 4.6** |
| **All 7 models** | | | | | | | |
| **+plan**<br>Reference high-level plan | 8 | 54 | 33.0 → 53.6 | 44.4 → 64.8 | 29 / 19 | 62.6 → 43.8 | 14.9 → 18.7 |
| **+state**<br>Plan + exact 3D state each turn | 8 | 54 | **33.0 → 87.8** | **44.4 → 94.4** | **48 / 2** | **62.6 → 12.0** | **14.9 → 3.0** |

n = (model, instance) pairs. ↑ / ↓ counts pairs with higher / lower scores; ties are omitted.

Pairs in which either run was abandoned after repeated API errors are excluded (Section 5).

‡ Evaluated on condition-specific instance subsets (Section 5; per-model breakdowns in Appendix C).

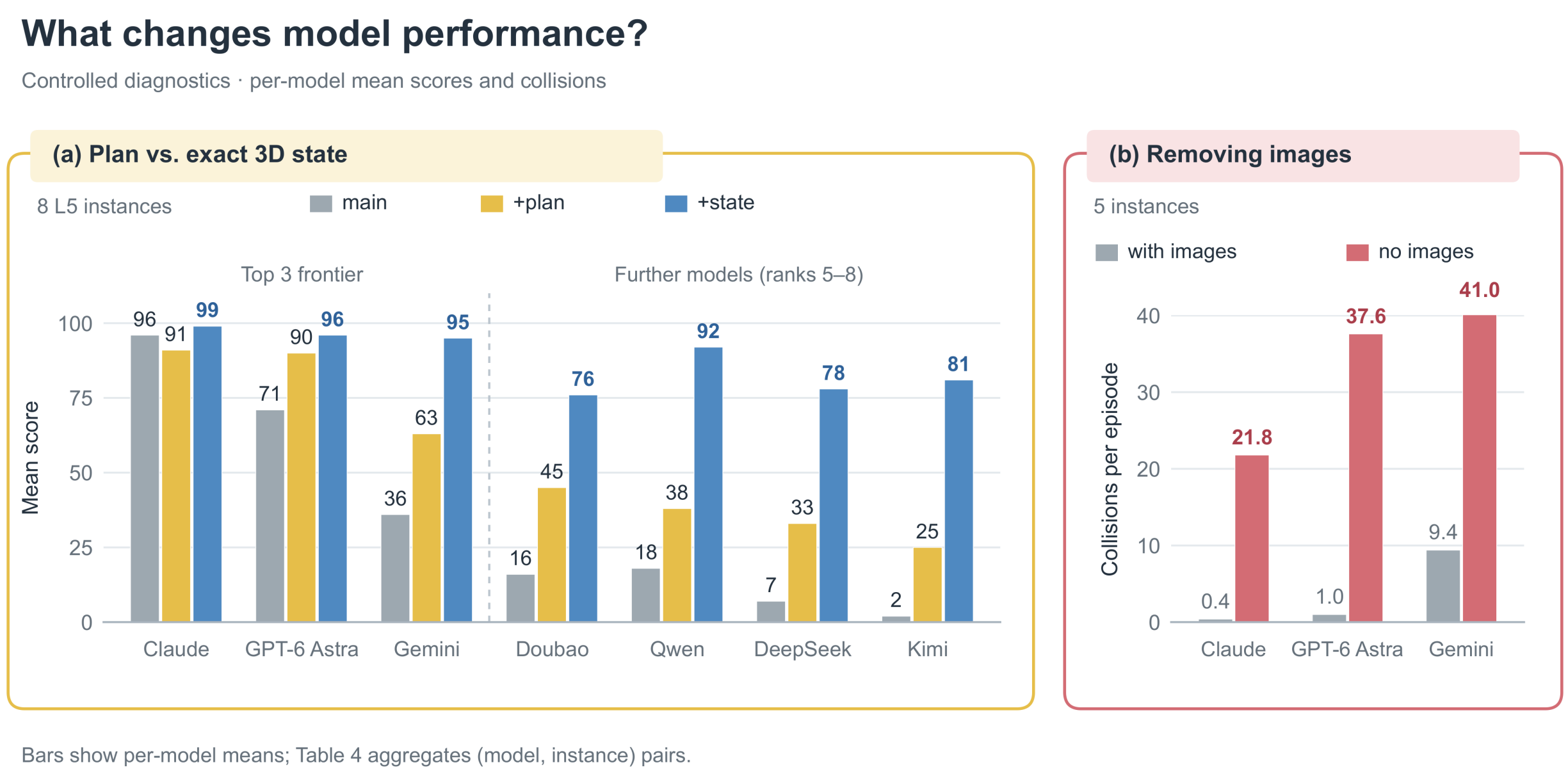

**Figure 7:** Controlled diagnostics per model. **(a)** Mean score under main, +plan, and +state on the eight shared L5 instances for the top-3 frontier models and the four further models; Claude Opus 5.5 is averaged over its six pairs with complete runs. **(b)** Collisions per episode with and without image input on the five blind-test instances (top-3 models).

**High-level planning ($\mathcal{P}_{\text{plan}}$) vs. 3D visual grounding ($\mathcal{G}_{\text{vis}}$).** Table 4 and Figure 7 summarize the diagnostics for the top-3 frontier models and four further models ranked 5th–8th in Table 3 (Doubao Seed 2.1 Pro, Qwen 3.8 Max, DeepSeek V4.1 Flash, and Kimi K3). Because +state supplies the reference plan as well as the exact 3D state, the two conditions are nested: +plan measures the value of the plan, and the step from +plan to +state measures the value of 3D state once the plan is known. For the top three (22 pairs on the eight shared L5 instances), the plan raises the mean score from 65.1 to 80.3 (closing 44% of the gap to 100; 10 pairs up, 8 down) but leaves collisions unchanged (5.2 → 5.2 per episode). Adding exact 3D state raises the score to 96.6 (closing 90%; SR 81.8% → 100.0%; 18 up, 0 down), shortens episodes sevenfold (36.7 → 5.3 turns), and cuts collisions to 0.5; it closes 83% of the gap that the plan leaves. The four further models average 10.9 points under main (SR 18.8%). The plan lifts them to 35.3 (SR 43.8%; 19 up, 11 down) while collisions rise from 21.5 to 28.0 per episode as agents push toward a known goal without accurate 3D coordinates; about 90% of this gain comes from the two discrete-search instances, the 3D gravity maze (P4) and the bevel-gear train (D8), where all four reach 100 once the plan is given, and the other six instances gain only 3.1 points per pair. The plan plus 3D state lifts them to 81.8 (SR 90.6%; 30 up, 2 down) and cuts collisions to 4.6. Across all 7 models (54 pairs), +plan closes 31% of the gap (33.0 → 53.6) and +state 82% (33.0 → 87.8; 48 up, 2 down); on the same runs, +state scores higher than +plan in 34 pairs, ties in 17 (16 of them at 100), and is lower in 3. The plan matters, mainly on discrete search, but once it is known, most of the remaining gap and nearly all collisions trace to 3D visual grounding ($\mathcal{G}_{\text{vis}}$).

**Without images, models probe by collision.** Removing visual input lowers the score from 73.9 to 17.7 and SR from 86.7% to 53.3%, while collisions surge about ninefold (3.6 → 33.5 per episode), with 12 pairs down and 0 up. The damage depends on whether collision feedback reveals scene geometry. On the thin-pier gorge (B1), shape-through-wall (M1), and C-ring on wire (M3), blocked actions or rollbacks alone locate obstacle boundaries, and the blind SR remains 77.8% (88.9% with images): collisions effectively serve as a tactile probe. On Klotski (P2) and the dynamic counterweight balance (B6), the global state must be observed visually and cannot be inferred from individual collisions, and SR collapses from 83.3% to 16.7%.

**Occluded tasks need active viewpoint control; the note and the split view mainly curb collisions.** Locking the camera lowers the score on heavily occluded instances from 58.3 to 37.8 (SR 68.4% → 47.4%), with 14 of 19 pairs declining. Ablating SAPA's persistent note (nonote) across six L5 tasks changes the 7-model mean score only modestly (31.8 → 29.5; 11 pairs up, 17 down) but lengthens episodes (60.1 → 67.5 turns) and raises collisions (14.6 → 17.8); GPT-6 Astra and Gemini 4 Argon are essentially unaffected (63.2 → 63.4), Claude Opus 5.5 completed only two of its six nonote runs, too few to judge, and the drop falls on the four further models (10.5 → 6.7; Table C.3). Removing the left-panel magenta/blue contact snapshot from rollback images (nosplit) leaves the mean score nearly unchanged (31.8 → 30.3) but raises collisions per episode by 59% (14.6 → 23.2), with 20 pairs declining vs. 9 improving, so explicit visual contact localization mainly prevents repeated blind collisions.

**Single-factor levers and supplementary ablations.** When individual L7 levers are added to L5 instances one at a time (3–15 paired runs per lever), the largest mean score drops for the three strongest models come from track rotation +rot (−12.5), the half-scale blueprint +blueprint (−10.3), and ceiling occlusion +occ (−10.2), while SR remains high. A top-3 ablation of the single-action-per-turn limit (cap1) is reported in Appendix C.

**Takeaway.** A reference plan helps, mainly on discrete search, but leaves top-3 collisions unchanged (5.2 → 5.2) and raises them for the four further models (21.5 → 28.0); adding exact 3D state on top of the plan raises the top-3 score to 96.6 and the four further models to 81.8 and cuts collisions to 0.5 and 4.6 per episode. Removing images and locking the camera point the same way: once the plan is known, 3D visual grounding is the binding constraint.

### 6.5 Behavioral Analysis

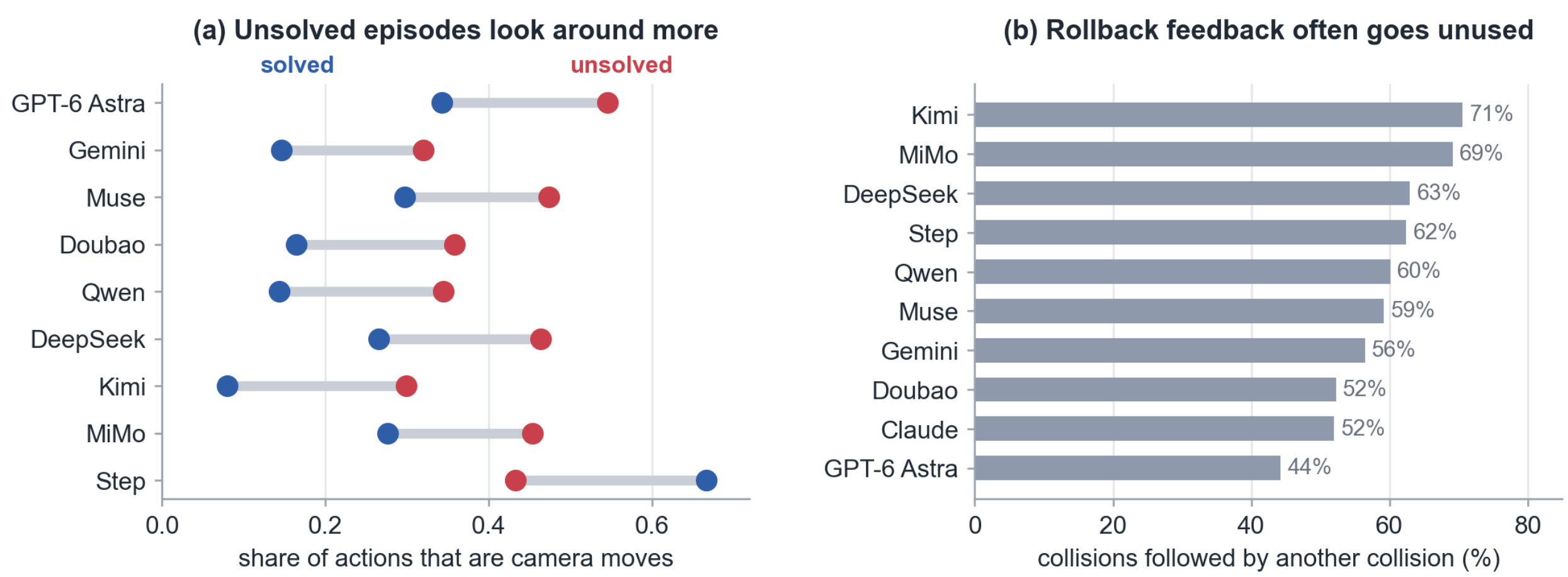

**Figure 8:** Behavioral signatures. **(a)** Share of actions that are camera moves in solved vs. unsolved episodes, for the nine models with both solved and unsolved trajectories. **(b)** Percentage of collisions that are immediately followed by another collision on the next turn.

We mine the per-turn trajectories of the 38 main-matrix instances that use the batched protocol (Figure 8; Appendix D). Claude Opus 5.5's three unsolved instances are the abandoned runs described in Section 5 and have no complete trajectory (they count as unsolved with score 0 in Table 3), so the statistics of unsolved trajectories in this section and Section 6.6 cover the other nine models.

**Failed episodes turn into observation loops.** For eight of the nine models with both solved and unsolved trajectories, the share of camera actions is higher in unsolved episodes than in solved ones (0.08–0.34 in solved vs. 0.30–0.55 in unsolved episodes); the exception, Step 5 Preview, solves a single batched instance. Failing agents keep searching for a viewpoint that will dispel their uncertainty instead of forming and testing hypotheses.

**Rollback feedback often goes unused.** After a batch is truncated, all 10 frontier models slightly shorten their next batch, and 16.5%–49.5% of next turns include a camera action. Yet 44%–71% of frontier collisions are followed by another collision on the very next turn; even GPT-6 Astra, the lowest of all models, repeats 44% of the time. The split screen explicitly marks the contact location, but models frequently fail to turn it into a corrected action.

**Batching saves calls.** Models output 1.29–2.52 actions per response, reducing model calls by 23%–60% (median 50%) relative to one call per action.

### 6.6 Failure-Mode Analysis

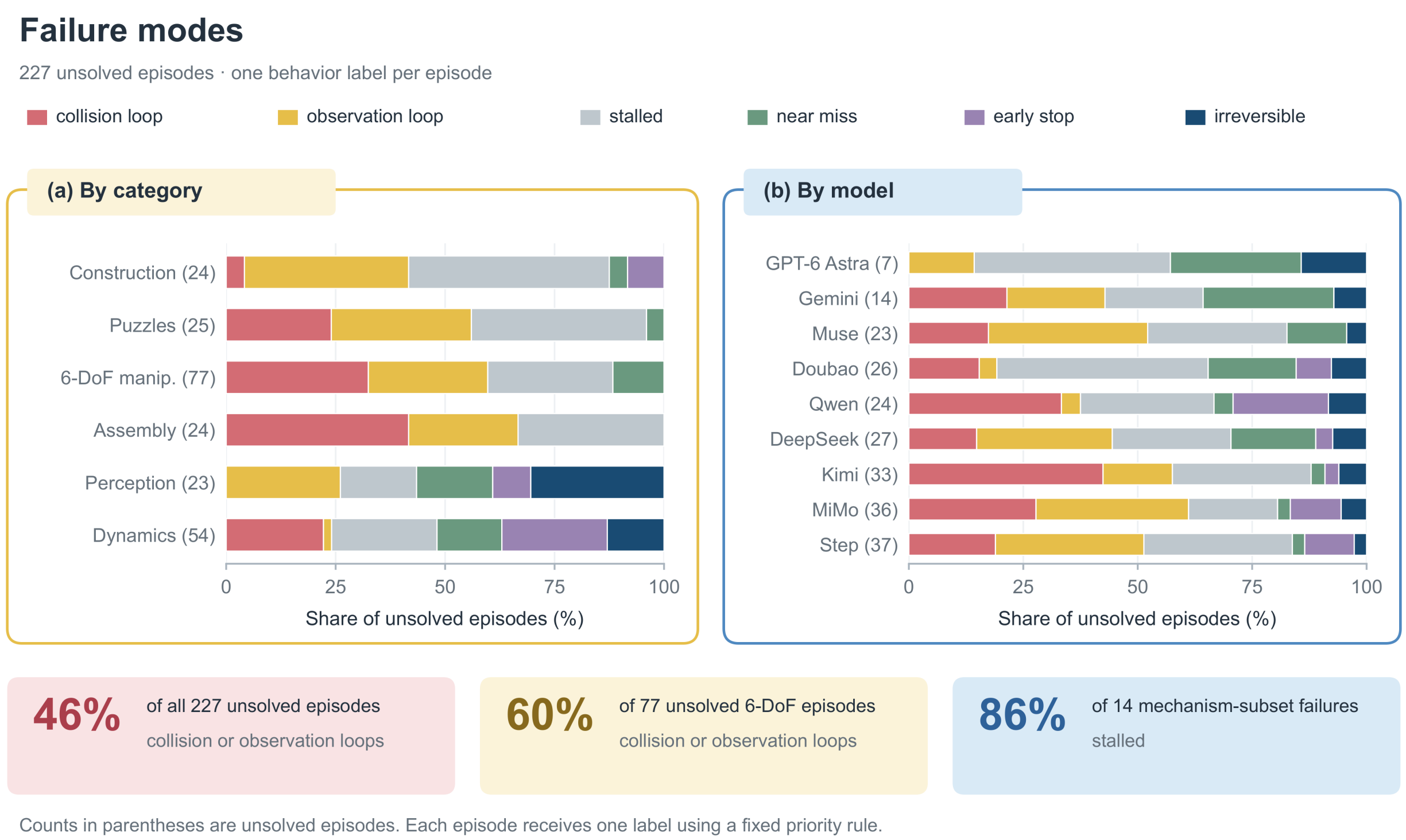

**Figure 9:** Failure modes of the 227 unsolved batched-protocol episodes with complete trajectories, broken down **(a)** by category and **(b)** by model (number of unsolved episodes in parentheses). Each episode receives a single signature in the priority order defined in Section 6.6.

To characterize failures at scale, we assign each of the 227 unsolved batched-protocol episodes a single behavioral signature, checked in the following order: *irreversible* (ending in a terminal failure, e.g., a marble entering the wrong cup or a wrong one-shot answer), *near miss* (maximum progress $\geq 0.75$), *collision loop* (collisions in at least 40% of turns), *observation loop* (at least 60% of actions are camera actions), *early stop* (the agent ends the episode voluntarily within 60 turns), and *stalled* (all remaining cases). These signatures are computed entirely from trajectories and are fully reproducible.

Under these thresholds, collision loops (24%) and observation loops (22%) together account for 46% of unsolved episodes (26%–62% under the other threshold combinations in our scripts); 30% are stalled, 10% are near misses, 7% stop early, and 6% end in an irreversible terminal failure, half of them wrong one-shot answers in perception tasks (Figure 9). In continuous 6-DoF manipulation, loops reach 60%: an agent that cannot localize accurately either hits the same obstacle repeatedly or keeps rotating the camera. 3D mechanism synthesis shows a different pattern: 86% of the 14 unsolved mechanism episodes (D8–D11) are stalled without loops, indicating that the difficulty there lies in mechanism design and topological planning rather than in local geometric grounding.

**Qualitative case: same instance, divergent visual grounding.** Figure 10 compares two models on the occluded helical wire task (D7-L5): a ring must travel along a wire hidden behind a screen, whose 3D course can be inferred only from an overhead shadow window, a rotated mirror side view, and collision rollbacks. Both models adopt the same high-level strategy, treating every blocked move as a probe that measures the wire's position. Claude Opus 5.5 converts these contact measurements into accurate geometric hypotheses and refines them step by step, solving the task in 34 turns with 15 collisions. Qwen 3.8 Max follows the same scheme, but two grounding errors, a reversed contact rule and an overhead slope read with the wrong sign, keep its hypothesis from converging: it records 93 collisions and stops at progress 0.46. With identical strategies, success is decided by the precision of 3D visual grounding.

**Takeaway.** Nearly half of unsolved agents are trapped in loops, hitting the same obstacle again and again or endlessly rotating the camera; most of the rest stall without a loop, especially in mechanism synthesis, where the difficulty lies in design and planning. Although the split-screen rollback marks the contact location, models still often fail to turn it into an effective geometric correction.

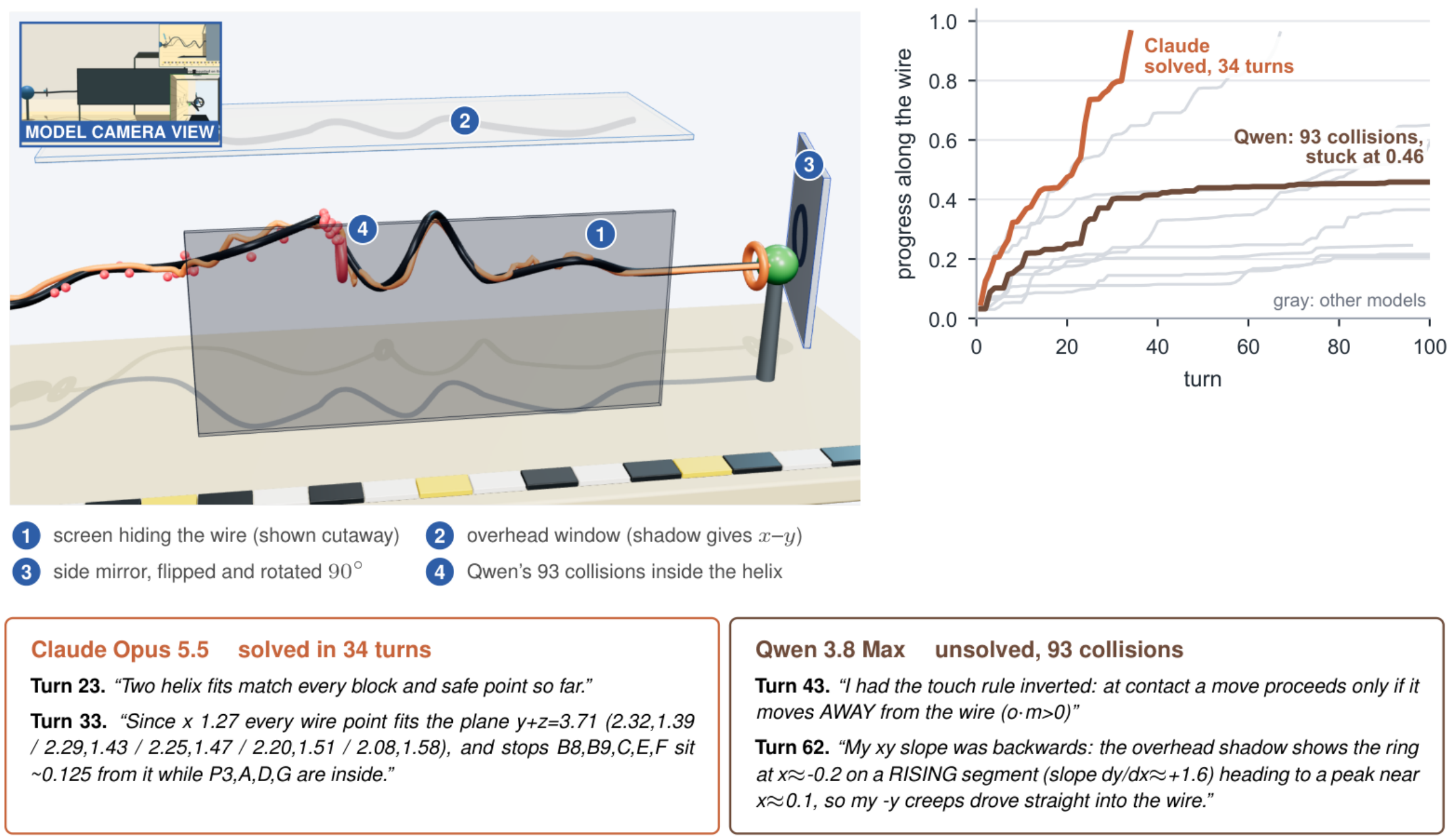

**Figure 10:** Case study on the occluded helical wire task (D7-L5). Left: replay of the scene. The ring must follow a wire hidden behind a screen (①, shown cutaway), whose course can be inferred only from the overhead shadow window (②, giving x–y), a flipped and rotated side mirror (③), and collision rollbacks; markers (④) show Qwen 3.8 Max's 93 collisions inside the helix. Top right: progress along the wire per turn (gray: other models). Bottom: excerpts of the models' reasoning. Claude Opus 5.5 turns contact measurements into a consistent geometric hypothesis and solves the instance in 34 turns; Qwen 3.8 Max's hypothesis does not converge after a reversed contact rule and a sign error in the overhead slope; it stops at progress 0.46.

## 6.7 Robustness

**Seeds.** For the three diagnostic models, rerunning six L5 families with seeds 101 and 102 and keeping the four on which every run completed yields a standard error of at most 2.0 points in each model's mean score, and the model ranking is identical across all three seeds (Appendix E).

**Rule perturbations.** Changing one rule of a classic puzzle leaves the number of frontier models that solve it essentially unchanged (P2: 7/10 vs. P3: 6/10; Appendix E), and an auxiliary Tower of Hanoi check outside the benchmark is at ceiling under both rules (10/10 standard vs. 10/10 restricted-disk). Both checks are consistent with solutions that are planned rather than recalled, but they cannot rule out memorization.

## 7 Discussion and Limitations

**The perception–control bottleneck from vision to continuous action.** Across the six categories, difficulty levers, controlled diagnostics, and per-turn trajectory analysis, the results draw a clear capability boundary for current frontier MLLMs. Temporal physical probing ($\mathcal{I}_{\text{phys}}$) holds up comparatively well: models adapt to the undisclosed physical changes of the two temporal L7 tiers (D3, D5) through strobe observation and interactive probing. A reference plan ($\mathcal{P}_{\text{plan}}$) helps, mostly on discrete search, but with the plan alone the three strongest models still collide as often as without it, whereas with the plan and exact 3D state they solve every diagnostic instance in 5.3 turns on average with 0.5 collisions per episode. The real bottleneck is the closed loop from raw pixels to continuous 3D action, that is, the coupling of metric visual grounding ($\mathcal{G}_{\text{vis}}$) with frame-aware control and rollback correction ($\mathcal{E}_{\text{ctrl}}$). Without images, models degrade into blind probing by collision; the L7 levers that hide geometry (+occ, +blueprint) or rotate the working frame (+rot) cause the steepest drops; and observation and collision loops form the largest group of unsolved episodes (46%). For MLLMs serving as planners in embodied physical systems, the most pressing direction is to build reliable metric 3D grounding from active multi-view observation and to convert collision and rollback feedback into calibrated spatial corrections.

**Scope.** The main matrix uses seed 100 and the hardest tier of each family; reruns with seeds 101 and 102 on the four L5 families with complete runs reproduce the ranking of the three diagnostic models, while the other seven models were run once. State injection (48 of 54 pairs up, 2 down) and image removal (12 of the 12 graded pairs down) move almost every pair in the same direction; the plan and L7 effects are clear on average but mixed at the pair level (+plan: 29 up, 19 down; L7: 37 down, 20 up). The procedural generators and L7 levers allow SpaceAct to scale to higher difficulty as models improve, and the rigid-body simulation with calibrated statics provides a controlled foundation for research on transfer to real robots.

## 8 Conclusion

We presented SpaceAct, an interactive benchmark that evaluates 3D spatial and temporal reasoning through closed-loop physical action and active perception. It contains 36 task families and 108 verified instances spanning structural construction, combinatorial puzzles, continuous 6-DoF manipulation, assembly, multi-view perception, and temporal and mechanism reasoning. The accompanying SAPA agent framework provides no privileged coordinates, enforces continuous collision and rigid-body statics constraints, freezes the clock during model deliberation, and offers controlled diagnostic modes that target visual grounding, physical identification, multi-step planning, and closed-loop control. Across 10 frontier MLLMs, SpaceAct yields strong separation and shows that, even with a correct high-level plan, the core bottleneck lies in 3D visual grounding and closed-loop control: for the three strongest models, a reference plan closes 44% of the remaining score gap and leaves collisions unchanged, whereas adding exact 3D state closes 90% and reduces collisions from 5.2 to 0.5, and collision and observation loops form the largest group of unsolved episodes. The same ordering holds for four further models, whose mean score rises from 10.9 to 35.3 with the plan and to 81.8 with the plan and exact 3D state. SpaceAct and SAPA provide a controlled, diagnosable testbed for multimodal agents that perceive, reason, and act in the 3D physical world.

## Appendix A  Task Family Catalog and Verification Details

Figure A.1 shows the initial observation of every task family as rendered to the model. Table A.1 summarizes the 36 task families and 108 verified instances across the six categories of SpaceAct, listing the identifier, difficulty tiers, primary action primitives, physical or geometric constraints, and deterministic verification method.

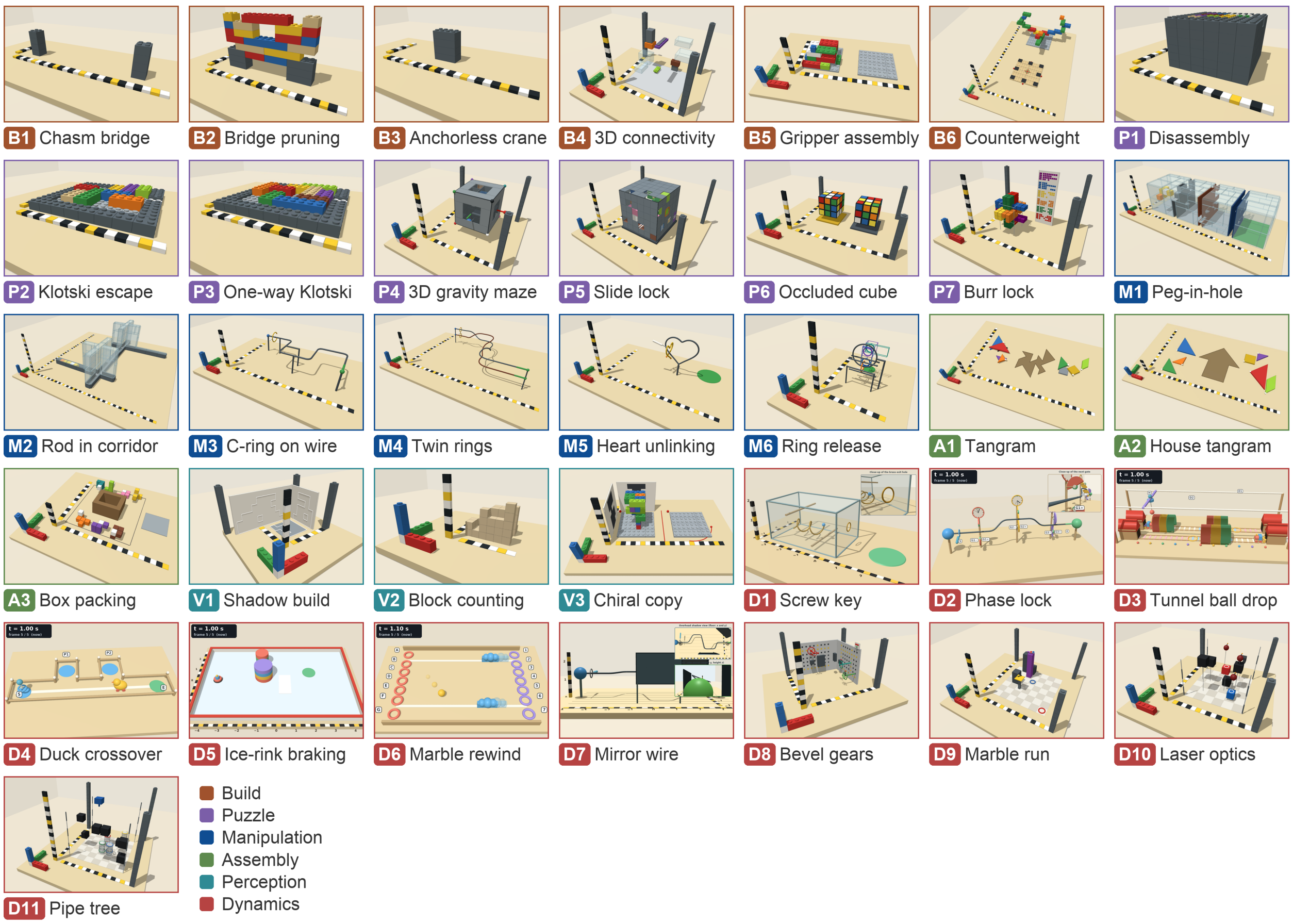

**Figure A.1:** Initial observation of each SpaceAct task family, as rendered to the model at the start of an episode (one representative instance per family). Labels follow Table A.1, and colors mark the six categories. Dynamic tasks show the last of their five strobe frames; some tasks include in-scene auxiliary views, such as the overhead shadow window and side mirror in D7.

**Table A.1:** Overview of the 36 SpaceAct task families across the six categories.

| ID | Task family | Cat. | Tiers (#Inst.) | Primary primitives & constraints | Reference solver & shortcut filter |
|---|---|---|---|---|---|
| B1 | Slender-pier chasm bridge | Build | Fixed + 2 variants (3) | place/remove, submit; 16-stud span, 3-zone load ≥ 1800g | StableLego-calibrated RBE LP solver |
| B2 | Load-bearing bridge pruning | Build | Fixed + 2 variants (3) | remove, submit; prune 18-brick bridge to ≤ 14.8g, load ≥ 2500g | RBE LP solver + collapse rollback |
| B3 | Anchorless tower crane | Build | Fixed + 2 variants (3) | place/remove, submit; step-by-step tipping equilibrium, hook ≥ 110g | Stepwise RBE moment equilibrium |
| B4 | 3D anchor connectivity | Build | L5 (3) | place/remove; stud-to-tube only, overhang ≤ 3 studs, L5 heterochromatic | 3D shortest-tree constructive solver |
| B5 | Two-finger gripper assembly | Build | L5 (3) | place/remove; side-finger clearance, temporary gray jig brick | Topological assembly-order search |
| B6 | Dynamic counterweight balance | Build | L5 (3) | place/remove; floating plate CoM $\lvert\Delta x\rvert,\lvert\Delta y\rvert \le \tau$ at every step | BFS over (model bricks, 4x weights) |
| P1 | 6-axis occluded disassembly | Puzzle | L5 (3) | pull(brick, dir); up to 20 enclosing bricks & 2 buried targets | Exact dependency-graph BFS (12/14/14) |
| P2 | Klotski traffic escape | Puzzle | Fixed + 2 variants (3) | slide(car, dir, steps) | Exact state-space BFS (10/17/16 moves) |

| ID | Task family | Cat. | Tiers (#Inst.) | Primary primitives & constraints | Reference solver & shortcut filter |
|---|---|---|---|---|---|
| P3 | Klotski with one-way vehicle | Puzzle | Fixed + 2 variants (3) | slide(car, dir, steps); one-way vehicle | Exact state-space BFS (12/20/18 moves) |
| P4 | 24-orientation 3D gravity maze | Puzzle | L5 (3) | rotate(axis, ±90°); 2 balls in L5 + opaque outer shell | SO(3) × ball-position BFS |
| P5 | 3-axis interlocking slide lock | Puzzle | L5 (3) | slide(bar, steps); 7–12 bars in $5^3/6^3$ frame | State-space BFS (depth ≥ 10 at L5) |
| P6 | Bottom-occluded 3×3×3 cube | Puzzle | L5 (3) | turn(face, ±90°); bottom face inferred from piece inventory | Group-search BFS + uniqueness check |
| P7 | Blueprint notched burr lock | Puzzle | L5 (3) | slide(bar, steps); notches on poster only, L5 push-then-retract | State-space BFS + non-monotone filter |
| M1 | Multi-wall chiral peg-in-hole | Manip. | L5, L7 ×2 (3) | move, rotate; 0.05-stud sweep, up to 3 walls + chiral decoys | Lattice SE(3) Dijkstra + lever map |
| M2 | Eccentric rod in glass corridor | Manip. | L5, L7 ×2 (3) | move, rotate; 1×1 roofed corridor, turning bay / vertical chimney | Lattice SE(3) Dijkstra + lever map |
| M3 | Slotted C-ring on 3D wire | Manip. | L5, L7 ×2 (3) | move, rotate; yaw at corners to align gap with standoffs | Swept-volume wire-frame planner |
| M4 | Twin rings on braided wire | Manip. | L5, L7 ×2 (3) | move, rotate; coupled roll along double-helix wire | Swept-volume helical planner |
| M5 | Interlocking heart unlinking | Manip. | L5, L7 ×2 (3) | move, rotate; 3D unlinking through skewed bottom gap | Multi-stage SE(3) unlinking planner |
| M6 | Borrowed-clearance ring release | Manip. | L5 (3) | grab, move, rotate; move 1–3 helper rings in limit boxes first | Multi-ring BFS + no-direct-exit filter |
| A1 | Perspective tangram fitting | Assem. | L5, L7 ×2 (3) | pick, rotate, flip, place; el ≤ 60°, decoys, rotated/poster tray | Exact polygon cover + perspective check |
| A2 | Perspective tangram (house target) | Assem. | L3 (3) | pick, rotate, flip, place; el ≤ 60°, decoys, rotated/poster tray | Exact polygon cover + perspective check |
| A3 | Opaque-box 3D polycube packing | Assem. | L5 (3) | pick, rotate, drop; el ≤ 60°, occluded lower layers + chiral decoy | 3D exact cover (DLX) + chiral filter |
| V1 | Multi-wall & oblique shadow build | Percep. | L5 (3) | place, remove; N/E orthogonal + L5 45° oblique shadow windows | Constructive projection solver |
| V2 | Low-elevation block counting | Percep. | L5 (3) | camera(el ≤ 40°), answer(count); gravity column support inference | Visibility hull + unique-count check |
| V3 | Occluded chiral/rotated copy | Percep. | L5 (3) | place, remove; 180°/mirror/mirror+90° + back-wall silhouettes | Coordinate-transform reference builder |
| D1 | Helical screw key extraction | Dyn. | Fixed + L5 ×2 (3) | screw(dtheta, dx), pull; coupled pitch constraint | Analytical pitch-coupled solver |
| D2 | 3-gate rotational phase lock | Dyn. | L5 (3) | wait(dt), advance; 5-frame strobe angular velocity & phase | Continuous-time phase-window solver |
| D3 | Multi-zone tunnel ball drop | Dyn. | L5, L7 ×2 (3) | wait(dt), drop; 3 speed zones, L5 dual tracks, L7 hidden speed | Piecewise-kinematic lead-time solver |
| D4 | Dual-track duck crossover | Dyn. | L5 (3) | move, wait; slow puddles + L5 bidirectional tracks & crossovers | Space-time collision-free BFS |
| D5 | 2D ice-rink momentum braking | Dyn. | L5, L7 ×2 (3) | thrust(power, dir, dt); block towers, L5 snow patch, L7 tilt drift | 2D second-order trajectoryshooting |
| D6 | 7×7 marble-cloud rewind | Dyn. | L5 (3) | answer(origins); elastic 90° collisions + color swaps | Exact time-reversal simulator + filter |
| D7 | Shadow & rotated-mirror wire | Dyn. | L5 (3) | move(dx,dy,dz); camera locked, occluded helix + 90° side mirror | 3D polyline waypoint follower |
| D8 | Dual-wall bevel gear train | Dyn. | L5 (3) | place_gear, crank; N+E walls, bevel_16, visual speed board & gates | Gear-graph ratio & phase solver |
| D9 | Timed dual-marble gravity run | Dyn. | L5 (3) | place_track, release; L3 pedal gate relay, L5 flip-flop switch | Discrete-event marble timing simulator |
| D10 | Equal-path 3D laser optics | Dyn. | L5 (3) | place_mirror/prism, fire; 3D tilt mirrors, L5 simultaneous arrival | 3D ray-tracing path-length solver |
| D11 | Overflow & splitter pipe tree | Dyn. | L5 (3) | place_pipe, open_valve; cup color rings, ÷2 splitters & overflow tees | Flow-conservation tree solver |

Figure A.2 shows, at full resolution, the two kinds of images that Figure 4(c) depicts in miniature: a five-frame strobe observation and a split-screen rollback view.

**(a)** Five-frame strobe observation of a dynamic task (D3-L5, GPT-6 Astra)

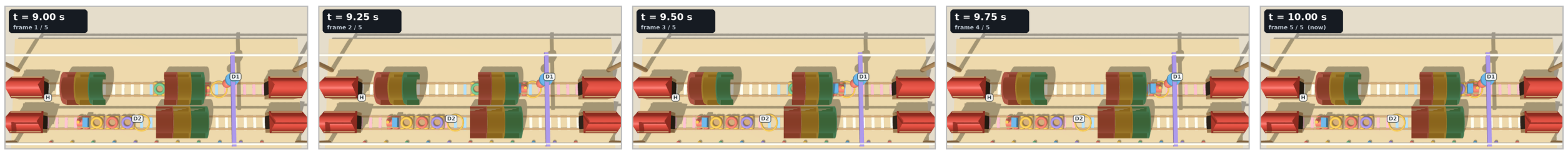

**(b)** Split-screen rollback feedback (B1, GPT-6 Astra)

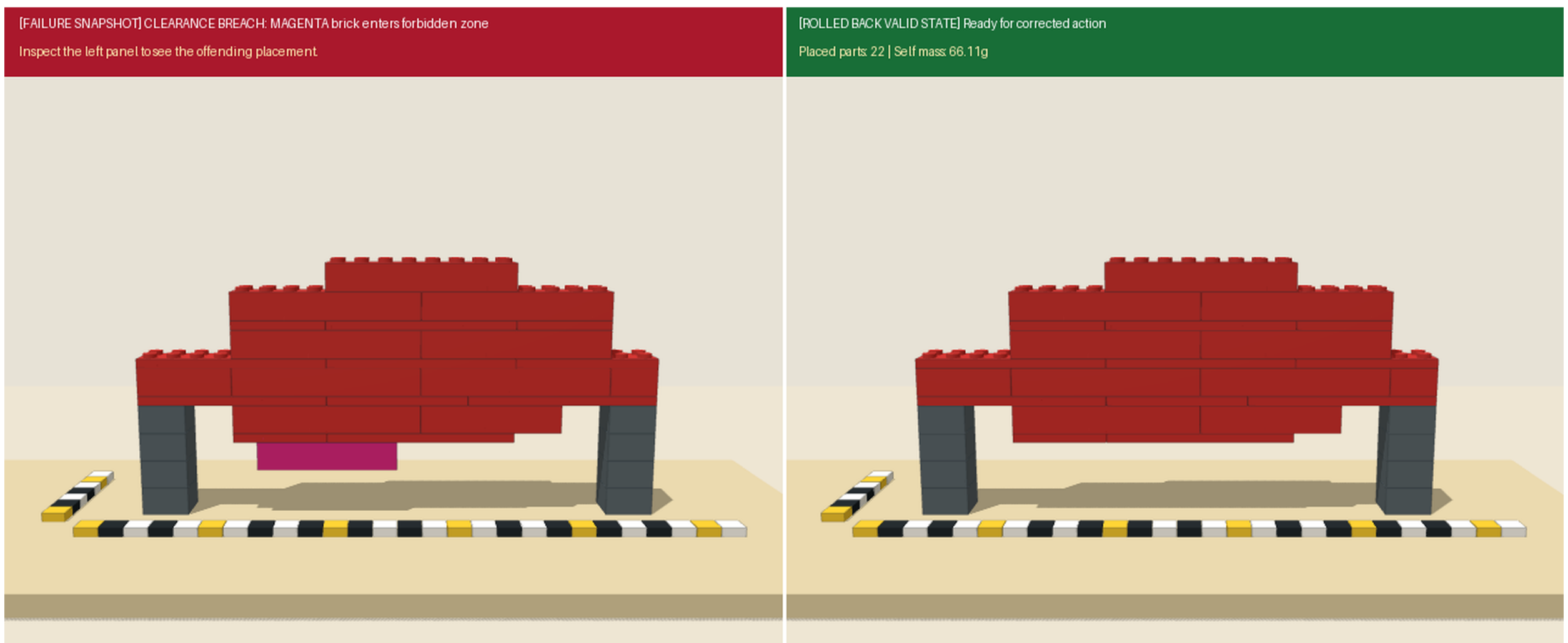

Failure snapshot: offending brick in magenta

Rolled back to the last valid state

**Figure A.2:** Observation and feedback images at full resolution. **(a)** Five-frame strobe observation from a dynamic task (D3-L5, GPT-6 Astra): frames from the same viewpoint at 0.25-s intervals (t = 9.00–10.00 s), each stamped with its time; the clock stays frozen while the model deliberates. **(b)** 1280 × 640 split-screen rollback view (B1, GPT-6 Astra): the left half shows the moment a placed brick (magenta) enters the forbidden clearance zone, and the right half shows the valid state after rollback, from which the next action starts. The labels below each half summarize the banners that the model receives.

## Appendix B Evaluated Model Configurations

Table B.1 lists the 10 evaluated frontier general and reasoning MLLMs, all accessed via public APIs.

**Table B.1:** Configuration of the 10 frontier MLLMs.

| Model | Category | Reasoning configuration | Max output tokens |
|---|---|---|---|
| Claude Opus 5.5 | Frontier MLLM (API) | reasoning_effort = max (streaming) | 128,000 |
| GPT-6 Astra | Frontier MLLM (API) | reasoning_effort = xhigh | 128,000 |
| Gemini 4 Argon | Frontier MLLM (API) | thinkingLevel = HIGH | 65,536 |
| Muse Wonder 1.0 | Frontier MLLM (API) | reasoning_effort = max (Responses API) | 131,072 |
| Doubao Seed 2.1 Pro | Frontier MLLM (API) | reasoning_effort = max | 262,144 |
| Qwen 3.8 Max | Frontier MLLM (API) | reasoning_effort = max | 131,072 |
| DeepSeek V4.1 Flash | Frontier MLLM (API) | reasoning_effort = max | 393,216 |
| Kimi K3 | Frontier MLLM (API) | Native long-CoT thinking enabled | 131,072 |
| MiMo V2.6 Pro | Frontier MLLM (API) | Default thinking enabled | 131,072 |
| Step 5 Preview | Frontier MLLM (API) | reasoning_effort = high | 65,536 |

## Appendix C Supplementary Diagnostic and Ablation Studies

**C.1 Single-factor L7 difficulty levers.** Table C.1 reports the paired effect of adding each individual L7 perceptual or coordinate-frame lever to L5 instances across Claude Opus 5.5, GPT-6 Astra, and Gemini 4

Argon (16 lever–instance combinations, 48 episodes in total). Rotating the track off the stud grid (+rot), moving wire geometry into a half-scale orthographic blueprint (+blueprint), and adding opaque ceiling covers (+occ) cause the largest score drops (−10.2 to −12.5 points), whereas single-factor tool-frame control (+tool) increases the mean score on M5-L5 for GPT-6 Astra and Gemini 4 Argon by aligning rotations with the heart's own symmetry axis.

**Table C.1:** Single-factor L7 difficulty lever ablations on L5 instances (Claude Opus 5.5, GPT-6 Astra, Gemini 4 Argon).

| Single-factor lever | Tasks | *n* | Score (main → lever) | Δ Score | SR (%) | Turns | Coll. |
|---|---|---|---|---|---|---|---|
| +rot (rotated track / board) | M3, A1 | 6 | 84.9 → 72.4 | −12.5 | 100.0 → 100.0 | 11.7 → 18.7 | 2.5 → 6.2 |
| +blueprint (half-scale 2D views) | M3, M4 | 6 | 73.8 → 63.4 | −10.3 | 100.0 → 100.0 | 22.5 → 37.0 | 7.5 → 8.2 |
| +occ (opaque ceiling / cover) | M2, M3, A1 | 9 | 84.7 → 74.5 | −10.2 | 100.0 → 88.9 | 11.6 → 32.6 | 2.2 → 5.1 |
| +nonum (no distance feedback) | M1, M2, M3 | 9 | 75.8 → 70.7 | −5.1 | 88.9 → 88.9 | 29.3 → 36.0 | 2.9 → 11.0 |
| +poster (target on back poster) | A1 | 3 | 84.6 → 82.8 | −1.8 | 100.0 → 100.0 | 9.7 → 12.7 | 3.0 → 3.7 |
| +tool (payload-local axes) | M1–M5 | 15 | 64.4 → 69.9 | +5.6 | 80.0 → 100.0 | 38.1 → 22.9 | 6.9 → 5.9 |

**C.2 Single-action closed-loop (cap1) vs. batched open-loop (cap5) and SAPA component ablations.** Table C.2 compares default 1–5 action batching (cap5) against a 1-action-per-turn limit (cap1) across four spatial tasks (M1, M3, P4, A1) and three temporal tasks (D2, D3, D5), and Table C.3 reports the per-model breakdown of +plan, +state, nonote, and nosplit across all 7 diagnostic models.

**Table C.2:** Batched open-loop (cap5) vs. single-action closed-loop (cap1) control across L5 spatial and temporal tasks.

| Task subset | #Inst. | *n* | Score (cap5 → cap1) | SR (%) (cap5 → cap1) | Turns (cap5 → cap1) | Coll. (cap5 → cap1) |
|---|---|---|---|---|---|---|
| Spatial subset (M1, M3, P4, A1) | 4 | 10 | 60.6 → 63.9 | 70.0 → 70.0 | 44.1 → 51.8 | 3.00 → 2.40 |
| Temporal subset (D2, D3, D5) | 3 | 9 | 69.9 → 70.4 | 100.0 → 88.9 | 7.9 → 8.8 | 0.56 → 0.44 |
| **All cap1 diagnostic tasks** | **7** | **19** | **65.0 → 67.0** | **84.2 → 78.9** | **26.9 → 31.4** | **1.84 → 1.47** |

**Table C.3:** Per-model breakdown of +plan, +state, nonote, and nosplit diagnostics across the 7 diagnostic models (score / SR). Pairs with an abandoned run are excluded (Section 5). †Claude Opus 5.5: six tasks for main, +plan, and +state, and two tasks for nonote and nosplit.

| Model | main (8 tasks) | +plan (8 tasks) | +state (8 tasks) | nonote (main → w/o note) | nosplit (main → w/o split) |
|---|---|---|---|---|---|
| Claude Opus 5.5† | 95.7 / 100.0% | 91.2 / 100.0% | 98.8 / 100.0% | 100.0 → 98.9 | 100.0 → 100.0 |
| GPT-6 Astra | 70.8 / 87.5% | 89.9 / 100.0% | 96.2 / 100.0% | 81.1 → 82.9 | 81.1 → 83.1 |
| Gemini 4 Argon | 36.3 / 62.5% | 62.6 / 87.5% | 95.3 / 100.0% | 45.2 → 43.9 | 45.2 → 43.8 |
| Doubao Seed 2.1 Pro | 15.9 / 25.0% | 44.8 / 62.5% | 76.0 / 87.5% | 11.0 → 7.2 | 11.0 → 14.6 |
| Qwen 3.8 Max | 18.4 / 37.5% | 37.7 / 50.0% | 92.2 / 100.0% | 22.4 → 18.7 | 22.4 → 6.8 |
| DeepSeek V4.1 Flash | 7.1 / 12.5% | 33.5 / 37.5% | 78.4 / 87.5% | 7.8 → 1.0 | 7.8 → 10.4 |
| Kimi K3 | 2.1 / 0.0% | 25.1 / 25.0% | 80.5 / 87.5% | 0.7 → 0.1 | 0.7 → 0.1 |
| **4 Further Models Mean** | **10.9 / 18.8%** | **35.3 / 43.8%** | **81.8 / 90.6%** | **10.5 → 6.7** | **10.5 → 8.0** |
| **All 7 Models (pair mean)** | **33.0 / 44.4%** | **53.6 / 64.8%** | **87.8 / 94.4%** | **31.8 → 29.5** | **31.8 → 30.3** |

## Appendix D Per-Model Trajectory and Behavioral Signatures

Table D.1 reports per-model trajectory signatures across the 38 main-matrix instances that use the batched protocol: mean actions per response (k), percentage of model calls saved by batching, share of actions that are camera moves in solved vs. unsolved episodes, percentage of collision turns immediately followed by another collision on the next turn, percentage of collision turns followed by a camera inspection, mean batch size before vs. after a collision, and share of turns that update the persistent note.

**Table D.1:** Per-model behavioral metrics across the 38 batched-protocol tasks in the main matrix. Claude Opus 5.5 has no unsolved trajectory because its three unsolved runs were abandoned (Section 5).

| Model | Mean $k$ | Call save (%) | Cam. solved (%) | Cam. unsol. (%) | Repeat hit (%) | Cam. after hit (%) | $k$ before → after hit | Note use (%) |
|---|---|---|---|---|---|---|---|---|
| Claude Opus 5.5 | 2.35 | 57.5 | 26.9 | — | 52.0 | 44.0 | 3.46 → 3.24 | 90.3 |
| GPT-6 Astra | 1.63 | 38.7 | 34.3 | 54.6 | 44.2 | 49.5 | 2.45 → 1.97 | 87.7 |
| Gemini 4 Argon | 2.52 | 60.3 | 14.7 | 32.0 | 56.5 | 28.6 | 3.59 → 3.11 | 99.2 |
| Muse Wonder 1.0 | 1.84 | 45.6 | 29.7 | 47.4 | 59.2 | 32.1 | 3.37 → 2.98 | 99.9 |
| Doubao Seed 2.1 Pro | 1.88 | 46.8 | 16.5 | 35.8 | 52.2 | 29.8 | 2.64 → 2.31 | 98.3 |
| Qwen 3.8 Max | 2.49 | 59.8 | 14.3 | 34.4 | 60.1 | 31.8 | 3.62 → 3.04 | 99.4 |
| DeepSeek V4.1 Flash | 2.16 | 53.7 | 26.6 | 46.4 | 63.0 | 34.1 | 3.65 → 3.24 | 96.0 |
| Kimi K3 | 2.30 | 56.6 | 8.0 | 29.9 | 70.6 | 18.0 | 3.13 → 2.85 | 99.7 |
| MiMo V2.6 Pro | 1.29 | 22.7 | 27.6 | 45.4 | 69.1 | 16.5 | 1.92 → 1.77 | 73.2 |
| Step 5 Preview | 1.60 | 37.4 | 66.7 | 43.3 | 62.4 | 34.8 | 2.47 → 2.09 | 96.6 |

## Appendix E Robustness Across Random Seeds and Rule Perturbations

Table E.1 reports cross-seed stability across seeds 100, 101, and 102 for the three diagnostic models on the four L5 families (M1, M3, A1, B4) on which every run completed. P4 and A3 were also rerun, but Claude Opus 5.5's runs on them were abandoned at one or more seeds (Section 5), so both families are left out for all three models. Standard errors across seeds are at most 1.98 points and model rankings are invariant across all three seeds. Table E.2 compares standard and rule-perturbed puzzle variants across all 10 frontier models.

**Table E.1:** Cross-seed stability across seeds 100, 101, and 102 on four L5 families (M1, M3, A1, B4).

| Model | Seed 100 (Score / SR) | Seed 101 (Score / SR) | Seed 102 (Score / SR) | 3-Seed Mean ± SE |
|---|---|---|---|---|
| Claude Opus 5.5 | 98.7 / 100.0% | 99.5 / 100.0% | 96.9 / 100.0% | 98.4 ± 0.77 |
| GPT-6 Astra | 85.6 / 100.0% | 84.6 / 100.0% | 81.2 / 100.0% | 83.8 ± 1.33 |
| Gemini 4 Argon | 45.8 / 75.0% | 52.6 / 75.0% | 48.4 / 100.0% | 48.9 ± 1.98 |

**Table E.2:** Rule-perturbed puzzle variants vs. standard variants across the 10 frontier MLLMs.

| Puzzle pair | Rule perturbation | Standard solved (of 10) | Perturbed solved (of 10) |
|---|---|---|---|
| P2 vs. P3 | One-way vehicle forbidding backward clearing move (10 → 12 optimal moves) | 7 / 10 (70.0%) | 6 / 10 (60.0%) |
| Tower of Hanoi, standard vs. restricted-disk (auxiliary, outside the benchmark) | 4-disk Tower of Hanoi: smallest disk forbidden from moving directly between Peg A and Peg C (15 → 17 optimal moves) | 10 / 10 (100.0%) | 10 / 10 (100.0%) |